%% file: main.tex
\documentclass[lettersize,journal]{IEEEtran}
\usepackage{balance}
\usepackage{amsfonts,amsmath,amssymb,amsthm}
\usepackage{graphicx}
\usepackage{epstopdf}
\usepackage{cite}
\usepackage{graphicx}
\usepackage{subfigure}
\graphicspath{{./fig/}}
\usepackage{algorithm,algorithmicx,algpseudocode}
\usepackage[nolist]{acronym}
\usepackage{enumitem}
\usepackage{xcolor}
\ifpdf
  \DeclareGraphicsExtensions{.eps,.pdf,.png,.jpg}
\else
  \DeclareGraphicsExtensions{.eps}
\fi
\usepackage{amsopn}

\input{acronyms.tex}

\DeclareMathOperator*{\minimize}{minimize~}
\DeclareMathOperator*{\st}{subject\,to~}
\newcommand{\R}{\mathbb R}
\newcommand{\mc}{\mathcal}
\newcommand{\tr}{^\textsf{T}}

\begin{document}
\title{Reactive Robot Navigation Using Quasi-conformal Mappings and Control Barrier Functions}
\author{Gennaro Notomista \IEEEmembership{Member, IEEE}, Gary P. T. Choi, and Matteo Saveriano \IEEEmembership{Senior Member, IEEE}
\thanks{Gennaro Notomista is with the Department of Electrical and Computer Engineering, University of Waterloo, Waterloo, ON, Canada. {\tt gennaro.notomista@uwaterloo.ca}}
\thanks{Gary P. T. Choi is with the Department of Mathematics, The Chinese University of Hong Kong, Hong Kong. {\tt ptchoi@cuhk.edu.hk}}
\thanks{Matteo Saveriano is with the Department of Industrial Engineering, Univesity of Trento, Trento, Italy. {\tt matteo.saveriano@unitn.it}}
\thanks{This work has been submitted to the IEEE for possible publication. Copyright may be transferred without notice, after which this version may no longer be accessible.}}

\maketitle

\begin{abstract}
  This paper presents a robot control algorithm suitable for safe reactive navigation tasks in cluttered environments. The proposed approach consists of transforming the robot workspace into the \emph{ball world}, an artificial representation where all obstacle regions are closed balls. Starting from a polyhedral representation of obstacles in the environment, obtained using exteroceptive sensor readings, a computationally efficient mapping to ball-shaped obstacles is constructed using quasi-conformal mappings and M\"obius transformations. The geometry of the ball world is amenable to provably safe navigation tasks achieved via control barrier functions employed to ensure collision-free robot motions with guarantees both on safety and on the absence of deadlocks. The performance of the proposed navigation algorithm is showcased and analyzed via extensive simulations and experiments performed using different types of robotic systems, including manipulators and mobile robots.
\end{abstract}


\section{Introduction}
\label{sec:introduction}

\IEEEPARstart{S}{afety} of dynamical systems is receiving increasingly more attention thanks to recently developed theoretical and computational tools that allow us to formulate several safety-critical controllers as convex optimization control policies. Applications include robot collision avoidance \cite{wang2017safety}, energy-aware systems \cite{notomista2020persistification}, and safe learning \cite{aswani2013provably,saveriano2019learning}. Safety is intended as the forward invariance property of a subset of the state space of the dynamical system describing the system. That is, a system is safe with respect to a set $\mc S$ if the trajectory of the state, $x(t)$, satisfies $x(t_0)\in\mc S \implies x(t)\in\mc S \quad\forall t\ge t_0$. This notion of safety has been employed to formulate problems of safe motion planning for autonomous systems \cite{petti2005safe}, navigation \cite{loizou2017navigation,thyri2020reactive}, and autonomy \cite{jha2018safe,notomista2020long}. More recently, safe reinforcement learning problems have been formulated leveraging such a control-theoretic notion of safety \cite{cheng2019end,emam2022safe,ohnishi2019barrier,notomista2023constrained}.

At the motion planning stage, safety constraints have been considered, which typically result in non-convex problem formulations (see, e.g., \cite{aswani2013provably,hsu2023safety}). The recently developed \acp{cbf} \cite{ames2019control} gained popularity also thanks to the low computational complexity of convex optimization formulations, as well as their ability to encode a large variety of safety specifications. Nevertheless, the low computational complexity comes at the cost of a reactive controller synthesis approach, as opposed to planning-like strategies (see, e.g., \cite{breeden2022predictive} for a recent work trying to unify reactive and predictive techniques). When such reactive controllers are employed, it has been shown that the presence of competing objectives (more specifically, stability and safety) may generate undesired and asymptotically stable equilibrium points \cite{reis2020control}. This phenomenon is particularly critical as undesirable asymptotically stable equilibria exist even in the presence of convex unsafe regions (e.g., obstacles).

\begin{figure}
	\centering
	\includegraphics[width=\linewidth]{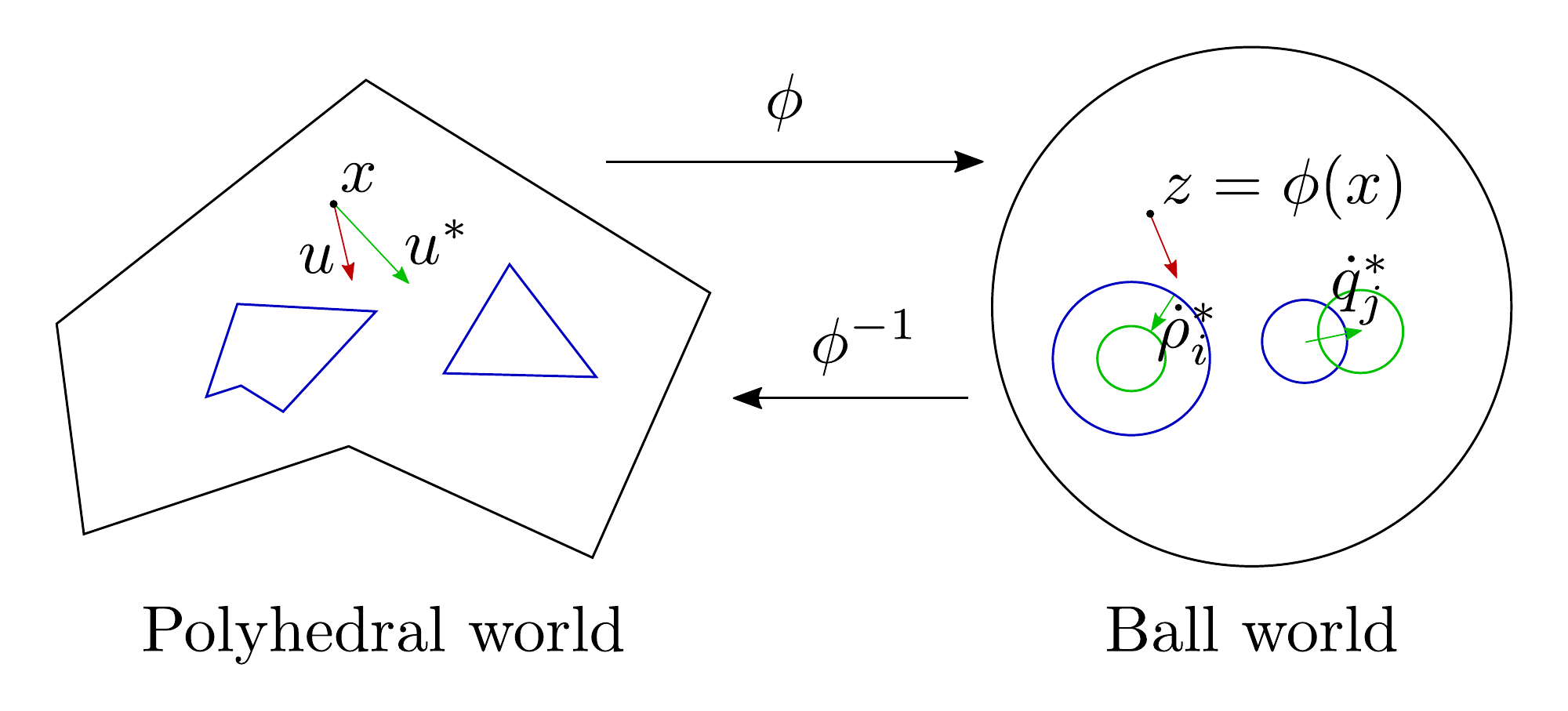}
	\caption{The approach proposed in this paper to ensure safety of dynamical systems in the presence of multiple non-convex unsafe regions consists of mapping the state space---the \textit{polyhedral world}---to a \textit{ball world}, where obstacles are closed balls or the complement of open balls. In the ball world, obstacles are transformed by changing their centers and radii, by means of the inputs $\dot\rho_i^*$ and $\dot q_j^*$. This lets the mapped state $z$, and hence the real state $x$, to remain safe.}
	\label{fig:proposedapproach}
\end{figure}

The presence of undesirable equilibrium points arises in robot navigation and path planning problems, where robots have to navigate around obstacles to reach a desired goal. Over the past several decades, different approaches for robot navigation and path planning via mathematical transformations have been proposed~\cite{rimon1990exact,rimon1991construction,rimon1992exact,belta2005discrete,rousseas2021harmonic} (see~\cite{goerzen2010survey} for a survey). For instance, Sarkar et al.~\cite{sarkar2009greedy} developed a method for greedy routing in sensor networks using Ricci flow. Loizou and Constantinou~\cite{loizou2017navigation,constantinou2020robot} tackled the navigation problem by mapping a star world to a domain called the point world. Vlantis et al.~\cite{vlantis2018robot} developed a method for robot navigation by transforming a real workspace into a punctured
disk using harmonic maps. Methods for motion planning using Schwarz--Christoffel mappings were proposed in \cite{gao2021conformal,notomista2018coverage}. In~\cite{fan2022robot}, Fan et al. proposed an iterative scheme for constructing conformal navigation transformations that map a complex workspace to a multiply-connected circle domain.

In our recent work \cite{notomista2021safety}, we presented an approach to mitigate the problem of undesirable asymptotically stable equilibria in the case of multiple non-convex unsafe regions. The control algorithm consists of mapping the system state space with possibly non-convex unsafe regions, to a space where unsafe regions are either closed balls or the complement of open balls. A safety-preserving controller is computed in the ball world and mapped, via an appropriately defined diffeomorphism, into the controller for the robot moving in the real world. The algorithm is formulated as a convex \ac{qp} regardless of the number and the convexity of unsafe regions. Figure~\ref{fig:proposedapproach} pictorially illustrates this idea.

In this paper, we design a computationally efficient reactive navigation policy for robotic systems which, thanks to the novel application of computational geometric tools to optimization-based robot control, significantly extends the range of applicability of the idea proposed in \cite{notomista2021safety}. More concretely, the proposed control algorithm improves on \cite{notomista2021safety} in the following ways:
\begin{enumerate}[label=(\roman*)]
	\item We employ \ac{qc} mappings to be able to represent obstacle regions in the real world---which will be referred to as the \textit{polyhedral world}---of which we do not have an analytical expression.
	\item We show how to compose partial \ac{qc} mappings in such a way to get a diffeomorphism between the polyhedral world and the ball world, where controllers are synthesized.
	\item We particularize the implementation of the developed algorithm for specific classes of systems modeling a large number of mobile robots employed in practical applications, namely, systems near-identity diffeomorphic to single integrators, feedback linearizable systems, and differentially flat systems.
	\item We report the results of extensive simulations and experiments with real robotic platforms to discuss the advantages and limitations of the proposed approach in terms of trade-offs between robustness and computational complexity.
\end{enumerate}
Compared to the work in \cite{notomista2021safety}, these contributions allow for the implementation of reactive controllers which, based on constrained-optimization-based formulations, guarantee a safe execution of robotic tasks, while preventing undesired equilibrium points from appearing and practically preventing the existence of deadlocks. Furthermore, we highlight how our proposed algorithm is amenable to be combined with existing tracking control techniques (based on, e.g., feedback linearization and differential flatness).

The remainder of the paper is organized as follows. In Section~\ref{sec:relatedwork}, we review the background and related work on robot navigation. In Section~\ref{sec:robotavoidance}, we describe our proposed safety-preserving algorithm leveraging \ac{qc} mappings and convex optimization problems. Numerical validation and comparisons are presented in Section~\ref{sec:validation}.
In Section~\ref{sec:robotic_applications}, we present both simulation and real experimental results for robotics applications. We conclude our work and discuss possible future directions in Section~\ref{sec:conclusions}.

\section{Background and Related Work}
\label{sec:relatedwork}

\subsection{Problem Definition}

In this paper, we consider robotic systems modeled by a control-affine dynamical system:
\begin{equation}
\dot x = f(x) + g(x) u,
\label{eq:nls}
\end{equation}
where $x\in\mathbb R^n$ is the robot state, $u\in\mathbb R^m$ is the robot control input, $f\colon\mathbb R^n\to\mathbb R^n$ and $g\colon\mathbb R^n\to\mathbb R^{n\times m}$ are locally Lipschitz continuous vector fields. The robots have to reach a desired goal in the environment, and this behavior is assumed to be achieved via a given control policy:
\[
\hat u \colon \mathbb R^n \times \mathbb R \to \mathbb R^m.
\]
This way,
\begin{equation}
\dot x = f(x) + g(x) \hat u(x,t),
\label{eq:diffflat}
\end{equation}
where $t$ denotes the time variable, makes the state $x$ evolve to reach a desired goal state.

A computationally efficient safety preserving control design for control affine dynamical systems is based on the use of \acp{cbf} \cite{ames2019control}, which results in a convex optimization control policy. The collision-free set---the \textit{safe} set---is defined as the zero-superlevel set of a continuously differentiable function $h$, i.e., the safe set $S$ is defined as $S = \{ x\in\mathbb R^n \colon h(x)\ge 0 \}$. Then, if it exists, the following controller guarantees collision-free motion of the robot:
\begin{equation}
	\begin{aligned}
		u^\star(x,t) = \arg\min_u &~\| u - \hat u(x,t) \|^2\\
		\mathrm{s.t.}&~L_f h(x) + L_g h(x) u \ge -\alpha(h(x)),
	\end{aligned}
	\label{eq:cbfqppoly}
\end{equation}
where $L_f h(x)=\frac{\partial h}{\partial x} f(x)$ and $L_g h(x)=\frac{\partial h}{\partial x} g(x)$ denote the Lie derivatives of $h$ in the direction of the vector fields $f$ and $g$, respectively, and $\alpha\colon\mathbb R\to\mathbb R$ is a class $\mathcal K$ function, i.e., a continuous, monotonically increasing function, with $\alpha(0)=0$.

For robot navigation tasks, the safe set $S$ is straightforwardly computable when the environment and the obstacles are balls. Therefore, the approach we take in this paper to devise a reactive robot navigation policy consists of the following steps:
\begin{enumerate}
	\item Reconstructing the polyhedral world from readings of exteroceptive sensors mounted on the robot.
	\item Mapping the polyhedral world to a ball world, where the workspace and the obstacles are closed balls.
	\item Solving for a robot control input that guarantees the collision-free motion of the robot in the ball world.
	\item Mapping the robot control input from the ball world to the polyhedral world and sending it to the robot.
\end{enumerate}
This is achieved by changing the position and size of the obstacles in the ball world, which in turn changes the diffeomorphism between the free space in the polyhedral world and the free space in the ball world. We then leverage this diffeomorphism and its Jacobian in order to compute the input for the robot in the polyhedral (real) world that ensures its collision-free motion without introducing undesired equilibrium points.

To this end, let $z$ denote the robot state in the ball world, and $\phi \colon \mathbb R^n \to \mathbb R^n$ the mapping from the polyhedral world to the ball world, namely $z = \phi(x) \in \mathbb R^n$ (see Fig.~\ref{fig:proposedapproach}). Assuming $\phi$ is continuously differentiable and invertible, we can write the robot dynamics in the ball world as follows:
\begin{equation}
	\label{eq:dynsysz}
	\begin{aligned}
		\dot z = \frac{\partial \phi}{\partial x} \dot x
		&= \frac{\partial \phi}{\partial x} f(x) + \frac{\partial \phi}{\partial x}g(x) u\\
		&= \underbrace{\frac{\partial \phi}{\partial x} (f\circ\phi^{-1})}_{=:f_z}(z) + \underbrace{\frac{\partial \phi}{\partial x} (g\circ\phi^{-1})}_{=:g_z}(z)u.
	\end{aligned}
\end{equation}
As the algorithm is implemented on real systems at discrete time stamps, we can relax the differentiability assumption and approximate derivatives with finite differences, by always keeping the theoretical guarantees of the \acp{cbf}. Having defined the \ac{cbf} $h$ whose zero-superlevel set is the collision-free set, we can solve for the robot controller $u^*$ as follows:
\begin{equation}
	\begin{aligned}
		u^\star(x,z,t) = \arg\min_u &~\| u - \hat u(x,t) \|^2\\
		\mathrm{s.t.}&~L_{f_z} h(z) + L_{g_z} h(z) u \ge -\alpha(h(z)).
	\end{aligned}
	\label{eq:cbfqpball}
\end{equation}
In the following section, we show how to build a diffeomorphism $\phi$ starting from a polyhedral description of the obstacles. The resulting function, together with its inverse and Jacobian, is locally Lipschitz continuous and bounded. As a result, the optimal solution to \eqref{eq:cbfqpball} is locally Lipschitz continuous \cite{ames2016control}.

\subsection{Building Mappings Between Real and Ball Worlds}
\label{subsec:analyticdiffeo}

In the ball world obstacles are mapped to convex sets (balls), which result in zero-measure sets of initial conditions from which deadlocks arise \cite{reis2020control}. More importantly, for the approach we propose in this paper, ball world obstacles can be parameterized by their position and radius. These two parameters will be controlled in our algorithm in order to ensure safe motion of the robot in the polyhedral world. In order to build a computationally tractable diffeomorphism between the polyhedral and the ball world, we leverage \ac{qc} mappings composed using the navigation functions in \cite{rimon1991construction}. In the following, we will briefly recall the definition of the latter for the case of star-shaped obstacles---which results in an analytic expression of the function---while the next section is devoted to the definition and properties of the \ac{qc} mappings, which allow us to obtain a mapping from the polyhedral world (where obstacles are polyhedrons) to the ball world.

Consider that the robot state is evolving in the $n$-dimensional manifold $\mc M^n$. In order to build an analytic diffeomorphism between the real world (containing star-shape obstacles) and the ball world, we will compose a number of real-valued analytic functions, denoted by $\beta_i \colon \mc M^n \to \R$, for which 0 is a regular value. The robot workspace will be denoted by $\mc W \subset \mc M^n$, a connected and compact $n$-dimensional submanifold of $\mc M^n$ such that
\begin{equation}
	\begin{aligned}
		\mc W^\circ &\subset \{ x\in\mc M^n \colon \beta_0(x) > 0 \},\\
		\partial \mc W &\subset \{ x\in\mc M^n \colon \beta_0(x) = 0 \}
	\end{aligned}
\end{equation}
are its interior and boundary, respectively. It is assumed that $M$ static obstacles are present in the workspace $\mc W$. As will be explained in the following, the mapping between the polyhedral and the ball world is updated at each iteration. Therefore, while the motion of the obstacles is not considered explicitly, it is accounted for implicitly. These are denoted by $\mc O_i$ (where $i = 1, 2, \dots, M$) and correspond to the interior of a connected and compact $n$-dimensional submanifold of $\mc M^n$ such that
\begin{equation}
	\begin{aligned}
		\bar {\mc O}_i &\subset \mc W^\circ \quad \forall i,\\
		\mc W \setminus \bar {\mc O}_i &\subset \{ x\in\mc M^n \colon \beta_i(x)>0 \},\\
		\partial \bar {\mc O}_i &\subset \{ x\in\mc M^n \colon \beta_i(x)=0 \},\\
		\bar {\mc O}_i &\cap \bar {\mc O}_j = \emptyset \quad \forall i\neq j.
	\end{aligned}
\end{equation}
With these definitions, the \textit{safe space} $\mc S$ (or free space) is given by
\begin{equation}
	\mc S = \mc W \setminus \bigcup\limits_{i=1}^M \mc O_i.
\end{equation}

In the ball world, both the robot workspace and the obstacles are balls. Then, an explicit representation of them can be as follows:
\begin{equation}
	\begin{aligned}
		\hat{\mc O}_0 &= \{ q \in\R^m \colon \underbrace{\rho_0^2 - \|q-q_0\|^2}_{\hat \beta_0(q)}<0 \},\\
		\hat{\mc O}_i &= \{ q \in\R^m \colon \underbrace{\|q-q_i\|^2 - \rho_i^2}_{\hat \beta_i(q)}<0 \},
	\end{aligned}
\end{equation}
where $\hat{\mc O}_0$ is the representation of the robot workspace $\mc W^\circ$, and $q_i$ and $\rho_i$, $i = 1, \ldots, M$, denote the center and the radius of the $i$-th obstacle, respectively. Thus, the safe space in the ball world is:
\begin{equation}
	\hat{\mc S} = \{ q \in \R^n \colon \hat\beta_0(q)\ge 0, \hat\beta_1(q)\ge 0, \ldots, \hat\beta_M(q)\ge 0\}.
\end{equation}
The functions $\beta_i$ are not easy to obtain in case of generic shapes of the obstacles. In the proposed approach, however, obstacles are conveniently mapped to balls, making the expression of such functions analytic and simple.

Following the procedure described in \cite{rimon1991construction}, we now describe how to obtain an analytic diffeomorphism between the real and the ball world in the case of star-shaped obstacles in the plane. The following is a parameterized example of such a diffeomorphism:
\begin{equation}
	\label{eq:combineddiffeo}
    \resizebox{\linewidth}{!}{$
	\phi_\lambda(x) = \sum_{i=0}^{M}\sigma_{i,\lambda}(x) \underbrace{\left(\rho_i b_i(x) + q_i\right)}_{\substack{\phi_i(x)~:=~i\text{-th}\\\text{obstacle diffeomorphism}}} + \sigma_{g,\lambda}(x) \left((x-x_g) + q_g)\right),$
 }
\end{equation}
where
\begin{equation}
	\begin{aligned}
		b_i(x) = \frac{\|x-x_i\|}{r_i(\theta)}\begin{bmatrix}
			\cos\theta\\
			\sin\theta
		\end{bmatrix},
	\end{aligned}
\end{equation}
with $\theta = \angle(x-x_i)$, $x_i$ is the center of the obstacle, and $r_i$ is the function describing how the radius changes as a function of the angle $\theta$.
The functions $\sigma_{i,\lambda}$ are defined as follows:
\begin{equation}
	\begin{aligned}
		\sigma_{i,\lambda} &= \frac{\gamma_g\bar\beta_i}{\gamma_g\bar\beta_i+\lambda \beta_i}, & i=0, \ldots, M,\\
		\sigma_{g,\lambda} &=1-\sum_{i=0}^{M} \sigma_{i,\lambda},
	\end{aligned}
\end{equation}
with $\gamma_g = \|x-x_g\|^2$ and $\bar\beta_i = \prod\limits_{\substack{j=0\\j\neq i}}^{M} \beta_j$. Finally, $x_g$ and $q_g$ are goal points in the real and ball world, respectively, which in the context of the algorithm presented in this paper can be set to an arbitrary point in the safe space in the real and ball world, respectively. Notice that, given the structure of the analytic diffeomorphism $\phi$, the point $x_g$ is mapped to $q_g$. While this approach works well for obstacles that can be accurately described by star-shaped sets, obtaining a mapping for obstacles described by generic polyhedrons is challenging. In the next section, we will introduce \ac{qc} mappings and show how to employ them to construct a transformation from a polyhedral to a ball world.

\section{Safe Robot Navigation Algorithm}
\label{sec:robotavoidance}

\subsection{Full Quasi-conformal Mapping}
\label{subsec:fullQC}

In this section, we develop a fast method for mapping a polyhedral world to a ball world using conformal mappings and quasi-conformal mappings. Conformal mappings are angle-preserving mappings that have been commonly used for shape transformations. More specifically, let $\phi:\mathbb{C} \to \mathbb{C}$ be a map on the complex plane with $\phi(z) = u_{\phi}(x,y) + i \ v_{\phi}(x,y)$, where $z=x+iy$, $u_{\phi},v_{\phi}$ are real-valued functions, and $i$ is the imaginary number with $i^2 = -1$. $\phi$ is conformal if its derivative $\phi'(z)$ is nonzero everywhere and it satisfies the Cauchy--Riemann equations $\frac{\partial u_{\phi}}{\partial x} = \frac{\partial v_{\phi}}{\partial y}$ and $\frac{\partial u_{\phi}}{\partial y} = -\frac{\partial v_{\phi}}{\partial x}$. While conformal mappings can preserve the local geometry under the transformations, most of the existing conformal mapping algorithms require a considerable amount of computation and hence are not suitable for real-time robot navigation. Here, we consider a generalization of conformal mappings called the quasi-conformal (QC) mappings, which are planar homeomorphisms that map small circles to small ellipses with bounded eccentricity~\cite{lehto1973quasiconformal}. Mathematically, a \ac{qc} mapping $\phi: \Omega_1 \to \Omega_2$ from a planar domain $\Omega_1\subset \mathbb{C} $ to another planar domain $\Omega_2\subset \mathbb{C}$ is an orientation-preserving homeomorphism that satisfies the Beltrami equation 
$\frac{\partial \phi}{\partial \bar{z}} = \mu(z) \frac{\partial \phi}{\partial z}$ for some complex-valued function $\mu$ with $\|\mu\|_{\infty} < 1$. We propose a fast \ac{qc} mapping method based on a recent conformal parameterization algorithm~\cite{choi2020efficient} with modifications. Specifically, since the safety-preserving algorithm does not require the mapping in this step to be perfectly conformal, we can relax the conformality requirement of the algorithm in~\cite{choi2020efficient} by simplifying certain procedures, thereby producing a \ac{qc} mapping more efficiently.

Let $\mathcal{S} \subset \mathbb{C}$ be a planar polyhedral domain with $k$ polygonal holes, and denote the boundary as $\partial \mathcal{S} = \Gamma_0 - \Gamma_1 - \dots - \Gamma_k$, where $\Gamma_0$ is the outer boundary and $\Gamma_1, \dots, \Gamma_k$ are the inner boundaries. We first fill all $k$ holes and obtain a simply connected domain $\tilde{\mathcal{S}}$. We can then compute a disk harmonic map $\varphi:\tilde{\mathcal{S}} \to \mathbb{D}$ by solving the Laplace equation
\begin{equation} \label{eqt:laplace}
	\Delta \varphi = 0,
\end{equation}
subject to a circular boundary constraint $\varphi(\Gamma_0) = \partial\mathbb{D}$, where $\Delta$ is the Laplace--Beltrami operator defined on $\tilde{\mathcal{S}}$ discretized using the cotangent formulation~\cite{pinkall1993computing} (see~\cite{choi2015fast} for more details). Next, we remove the filled regions and transform the $k$ holes in $\varphi(\mathcal{S})$ into $k$ circles such that the center of each circle is the centroid of the corresponding hole and the area of the circle equals that of the hole (here we remark that alternatively, one may also map the $k$ holes to $k$ circles based on some prescribed centers and radii). Finally, we obtain a \ac{qc} map $\widetilde{\varphi}:\mathcal{S}\to \mathbb{D}$ using the Linear Beltrami Solver (LBS) method~\cite{lui2013texture,choi2015flash} subject to the updated circular boundary constraints. More specifically, $\widetilde{\varphi}$ is obtained by solving the generalized Laplace equation
\begin{equation} \label{eqt:lbs}
	\nabla  \cdot (A \cdot \nabla \widetilde{\varphi}) = 0,
\end{equation}
subject to the updated boundary constraints for all $\Gamma_j$, where 
\begin{equation}
	A = \begin{pmatrix}
		\frac{(\operatorname{Re}(\mu_{\varphi})-1)^2 + (\operatorname{Im}(\mu_{\varphi}))^2}{1-|\mu_{\varphi}|^2} & -\frac{2\operatorname{Im}(\mu_{\varphi})}{1-|\mu_{\varphi}|^2}\\
		-\frac{2\operatorname{Im}(\mu_{\varphi})}{1-|\mu_{\varphi}|^2} & \frac{(\operatorname{Re}(\mu_{\varphi})+1)^2 +(\operatorname{Im}(\mu_{\varphi}))^2}{1-|\mu_{\varphi}|^2}
	\end{pmatrix}
\end{equation}
and $\mu_{\varphi} = \frac{\varphi_{\bar{z}}}{\varphi_z}$ is the Beltrami coefficient of the disk harmonic map $\varphi$. This completes the proposed fast method for mapping the polyhedral world to the ball world. In the discrete case, note that both Eq.~\eqref{eqt:laplace} and Eq.~\eqref{eqt:lbs} are $n_v \times n_v$ sparse linear systems, where $n_v$ is the number of vertices in the polyhedral domain $\mathcal{S}$. In other words, the Full QC method only requires solving two linear systems without any iterations and hence is highly efficient. Also, the resulting \ac{qc} map $\phi$ has bounded distortion as guaranteed by quasi-conformal theory.

\subsection{Partial Conformal Mapping}
\label{subsec:partialQC}

In the proposed method above, the entire polyhedral world is mapped to a ball world by solving linear systems. However, in some cases, it may be desirable to have an even more efficient method for mapping only part of the polyhedral world to a partial ball world, where the computation of the mapping should be as simple as possible for facilitating real-time navigation. Here, we develop an algorithm for conformally mapping a polygonal hole in the polyhedral world to the unit disk, and everything outside the hole to the exterior of the unit disk. To achieve this, we adopt a similar strategy as in~\cite{choi2020parallelizable} and utilize the geodesic algorithm~\cite{marshall2007convergence}, which consists of a series of analytic maps on the complex plane.

Denote the boundary vertices of the polygonal hole as $p_1, p_2, \dots, p_n$ (in anticlockwise order). We first consider the following mapping:
\begin{equation}
	\varphi_0(z) = \sqrt{\frac{z-p_2}{z - p_1}},
\end{equation}
with the branching $(-1)^{1/2} = i$, which maps $p_1$ to $\infty$, $p_2$ to 0, and the interior of the polygon to the right half-plane. We then construct a sequence of mappings $\varphi_1, \dots, \varphi_{n-2}$ such that the remaining $p_i$'s are mapped to the imaginary axis one by one via their compositions: 
\begin{equation}
	\varphi_{j}(z) = \sqrt{\left(\frac{\frac{\operatorname{Re}\xi_j }{|\xi_j|^2}z}{\frac{1+ \operatorname{Im}\xi_j }{|\xi_j|^2}zi}\right)^2-1},
\end{equation}
where $\xi_j = (\varphi_{j-1}\circ\varphi_{j-2} \circ \cdots \circ \varphi_0)(p_j)$, with the branching $(-1)^{1/2} = -i$. Next, we apply the following analytic map:
\begin{equation}
	\varphi_{n-1}(z) = \left(\frac{z}{1-\frac{z}{(\varphi_{n-2}\circ\varphi_{n-3} \circ \cdots \circ \varphi_0)(p_1)}}\right)^2,
\end{equation}
which enforces the interior of the polygon to be in the upper half plane while keeping $p_1$ to be mapped to $\infty$. The Cayley transform
\begin{equation}
	\varphi_n(z) = \frac{z - i}{z + i}
\end{equation}
can then be applied to map the real axis to the unit circle, the upper half plane to the interior of the unit disk, and the lower half plane to the exterior of the disk. Next, we normalize the transformation and ensure that $\infty$ remains fixed by applying a reflection with respect to the unit circle $\eta(z) = \frac{1}{\overline{z}}$, followed by a M\"obius transformation that shifts  $\infty$ to $0$:
\begin{equation}
	\tau(z) = \frac{z-(\eta \circ \varphi_n \circ \dots \circ \varphi_0)(\infty)}{1-\overline{(\eta \circ \varphi_n \circ \dots \circ \varphi_0)(\infty)}z}.
\end{equation}
Finally, we apply an inverse reflection $\eta^{-1}(z) = \frac{1}{\overline{z}} = \eta(z)$.

Altogether, the composition of the above-mentioned mappings $\Psi = \eta^{-1} \circ \tau \circ \eta \circ \varphi_n \circ \cdots \circ \varphi_0$ maps the polygonal hole to the unit disk and everything outside the hole to the exterior of the unit disk. Note that $\Psi$ is fully analytic and does not require any equation solving.

\subsection{Convex Optimization Formulation}
\label{subsec:QP}

In this section, we develop the main optimization-based controller, which is designed to move the positions of the obstacles as well as their radii so that the robot in the ball world moves in a collision-free fashion. Leveraging the diffeomorphism built using the techniques recalled in the previous section, the motion of the robot in the ball world is mapped into the motion of the robot in the polyhedral (real) world---see also the Appendix for more implementation details. It is important to remark that the motion of the obstacles in the ball world is artificial and does not correspond to the motion of the obstacles in the polyhedral world. While the approach proposed in this paper is capable of accounting for the latter, for sake of clarity we do not consider it in this work.

As the ball world is a mathematical representation of the real robot workspace, it can be freely deformed and modified in order to ensure the robot never collides with the obstacles present in the environment and with the boundary of the latter. This approach can be interpreted as a \textit{robot-avoidance paradigm}, as opposed to the more traditional obstacle-avoidance one. In these settings, obstacles in the ball world are displaced from their positions and shrunk with respect to their initial radii so that the following conditions are satisfied in the ball world:
\begin{enumerate}
	\item[(C1)] The robot does not collide with the obstacles.
	\item[(C2)] The obstacles do not collide with each other.
	\item[(C3)] The obstacles do not exit the environment.
	\item[(C4)] The obstacles do not overlap with the goal point $q_g$ used to build the mapping between real and ball worlds.
\end{enumerate}
Condition (C1) is directly related to safety: If the robot is kept in the safe set in the ball world, so is the case in the polyhedral world. Therefore, having the obstacles move away from the robot in the ball world will result in the robot avoiding the obstacles in the polyhedral world. Conditions (C2), (C3), and (C4) are introduced to always obtain a valid mapping between real and ball world by satisfying the assumptions introduced in the previous section. Given the simple geometry of the ball world, where every object is a ball, constraint-satisfaction for the obstacles can be efficiently enforced through the use of \acp{cbf}.

In this paper, we are going to enforce constraints on the obstacles by leveraging \acp{cbf}. In order to do so, we need to define the dynamics of the obstacle motion. We choose to control both the position of the center of each obstacle, $q_j$, via its velocity, $u_{q_j}$, as well as the radius of each obstacle, $\rho_j$, via the radius rate of change $u_{\rho_j}$. Therefore, we can define the following single integrator obstacle dynamical model:
\begin{equation}
	\begin{cases}
		\dot q_j = u_{q_j},\\
		\dot \rho_j = u_{\rho_j}.
	\end{cases}
\end{equation}
We are now ready to present the \acp{cbf} employed to enforce Conditions (C1), (C2), (C3), and (C4) introduced above.

\paragraph{CBF for Condition (C1)} The objective of keeping the robot at position $q$ away from obstacle $j$ can be enforced by defining formally the \ac{cbf} $h_j:=\hat\beta_j$, i.e.,
\begin{equation}
	h_j(q_j, \rho_j) = \| q_j - q \|^2 - \rho_j^2.
\end{equation}
The following inequality constraint on $u_{q_j}$ and $u_{\rho_j}$ can be defined in order to enforce the positivity of $h_j$ \cite{ames2019control}:
\begin{equation}
	\underbrace{\begin{bmatrix}
			-2(q_j-q)\tr & 2\rho_j
	\end{bmatrix}}_{=:A_{C1,j}}
	\begin{bmatrix}
		u_{q_j}\\
		u_{\rho_j}
	\end{bmatrix} \le \underbrace{-2(q_j-q)\tr \dot q + \alpha(h_j(q_j, \rho_j))}_{=:b_{C1,j}}.
\end{equation}
This constraint on the control inputs $u_{q_j}$ and $u_{\rho_j}$ will be enforced in an optimization program formulated to synthesize the controller to move and deform the obstacles in the ball world.

\paragraph{CBF for Condition (C2)} To make the obstacles not collide with each other, we can proceed to define the following \ac{cbf}:
\begin{equation}
	h_{jk}(q_j,q_k,\rho_j,\rho_k) = \| q_j - q_k \|^2 - (\rho_j+\rho_k)^2,
\end{equation}
and corresponding inequality constraint for the inputs to the obstacles:
\begin{equation}
    \resizebox{\linewidth}{!}{$
	\begin{aligned}
		&\underbrace{\begin{bmatrix}
				-2(q_j-q_k)\tr & 2(q_j-q_k)\tr & 2(\rho_j+\rho_k) & 2(\rho_j+\rho_k)
		\end{bmatrix}}_{=:A_{C2,jk}}
		\begin{bmatrix}
			u_{q_j}\\
			u_{q_k}\\
			u_{\rho_j}\\
			u_{\rho_k}
		\end{bmatrix}\\
		&\le \underbrace{\alpha(h_{jk}(q_j,q_k,\rho_j,\rho_k))}_{=:b_{C2,jk}}.
	\end{aligned}
 $}
\end{equation}

\paragraph{\ac{cbf} for Condition (C3)} To prevent the obstacle from exiting the environment, we can proceed similarly to the previous two cases and define the following \ac{cbf} for each obstacle $j$:
\begin{equation}
	h_{j0}(q_j,\rho_j) = (\rho_0-\rho_j)^2 - \| q_j - q_0 \|^2,
\end{equation}
resulting in the following constraint on $u_{q_j}$ and $u_{\rho_j}$:
\begin{equation}
	\underbrace{\begin{bmatrix}
			2(q_j-q_0)\tr & 2(\rho_0-\rho_j)
	\end{bmatrix}}_{=:A_{C3,j}}
	\begin{bmatrix}
		u_{q_j}\\
		u_{\rho_j}
	\end{bmatrix}\le \underbrace{\alpha(h_{j0}(q_j,\rho_j))}_{=:b_{C3,j}}.
\end{equation}

\paragraph{\ac{cbf} for Condition (C4)} Finally, the non-overlapping condition between any of the obstacles and the goal of the analytic diffeomorphism can be obtained with the following \ac{cbf}:
\begin{equation}
	h_{jg}(q_j,\rho_j) = \| q_j - q_g \|^2 - \rho_j^2,
\end{equation}
from which the following constraint on $u_{q_j}$ and $u_{\rho_j}$ follows:
\begin{equation}
	\underbrace{\begin{bmatrix}
			-2(q_j-q_g)\tr & 2\rho_j
	\end{bmatrix}}_{=:A_{C4,j}}
	\begin{bmatrix}
		u_{q_j}\\
		u_{\rho_j}
	\end{bmatrix} \le \underbrace{\alpha(h_{jg}(q_j,\rho_j))}_{=:b_{C4,j}}.
\end{equation}

At this point, in order to select the control input for the obstacles, we formulate the following \ac{qp}:
\begin{equation}
	\label{eq:mainqp}
	\begin{aligned}
		\minimize_{u_q,u_\rho} &\|u_q-\hat u_q\|^2 + \kappa \|u_\rho-\hat u_\rho\|^2\\
		\st & \begin{bmatrix}
			A_{C1,j}\\
			A_{C3,j}\\
			A_{C4,j}
		\end{bmatrix} \begin{bmatrix}
			u_{q_j}\\
			u_{\rho_j}
		\end{bmatrix}\le \begin{bmatrix}
			b_{C1,j}\\
			b_{C3,j}\\
			b_{C4,j}
		\end{bmatrix} \quad\forall j=1,\ldots, M\\
		& A_{C2,jk}
		\begin{bmatrix}
			u_{q_j}\\
			u_{q_k}\\
			u_{\rho_j}\\
			u_{\rho_k}
		\end{bmatrix}\le b_{C2,jk} \quad\forall j,k=1, \ldots, M,j>k,
	\end{aligned}
\end{equation}
where $u_q$ and $u_\rho$ are the stacked obstacle velocity and radius change rate, respectively,
\begin{equation}
	\begin{aligned}
		u_q&=\begin{bmatrix}
			u_{q_1}\tr & \ldots & u_{q_M}\tr
		\end{bmatrix}\tr,\\
		u_\rho&=\begin{bmatrix}
			u_{\rho_1} & \ldots & u_{\rho_M}
		\end{bmatrix}\tr,
	\end{aligned}
\end{equation}
and $\hat u_q$ and $\hat u_\rho$ are the stacked nominal control inputs for the velocity and the radius change rate of the obstacles defined as follows in order to keep the original positions $q_j(0)$ and radius $\rho_j(0)$ if possible:
\begin{equation}
	\label{eq:obsthat}
	\begin{aligned}
		\hat u_{q,j} &= K_p \left( q_j(0) - q_j(t) \right),\\
		\hat u_{\rho,j} &= K_p \left( \rho_j(0) - \rho_j(t) \right),
	\end{aligned}
\end{equation}
where $K_p>0$ is a controller gain. The parameter $\kappa$ determines the relative weight between the change of position and the change of radius of the obstacles, and can be chosen to prioritize one over the other. In \cite{notomista2021safety}, it is shown that the \ac{qp} in \eqref{eq:mainqp} is always feasible, regardless of the number of obstacles, their relative position to the robot and to the boundary of the environment.

\begin{algorithm}
	\caption{Safety with multiple obstacles}
	\label{alg:safemultipleconcave}
	\begin{algorithmic}[1]
		\Require $\phi$, $\Delta t$, $\alpha$, $\kappa$, $K_p$, $q_i(t_0)$ and $\rho_i(t_0)$, $i=0,\ldots,M$
		\State $k = 0$
		\While{true}
		\State $k \leftarrow k+1$
		\State $\dot q^{(k)} = L_f \phi^{(k)}\left(x^{(k)}\right) + L_g \phi^{(k)}\left(x^{(k)}\right)u^{(k)}$
		\State Compute $\hat u_q$ and $\hat u_\rho$ \Comment{\eqref{eq:obsthat}}
		\State Compute $u_q^*$ and $u_\rho^*$ \Comment{\eqref{eq:mainqp}}
		\State $q_i^{(k+1)} \leftarrow q_i^{(k)} + u_{qi}^*\Delta t,~i=0,\ldots,M$
		\State $\rho_i^{(k+1)} \leftarrow \rho_i^{(k)} + u_{\rho i}^*\Delta t,~i=0,\ldots,M$
		\State Update $\phi^{(k+1)}$
		\State $\dot x^{(k)} = \dfrac{\partial {\phi^{(k+1)}}^{-1}}{\partial q}\dot q^{(k)}$
		\State Compute $u^{(k+1)}$ to track $\dot x^{(k)}$ \Comment{$(\star)$}
		\State Apply $u^{(k+1)}$
		\EndWhile
	\end{algorithmic}
\end{algorithm}

Algorithm~\ref{alg:safemultipleconcave} summarizes the required steps to execute the safe robot navigation algorithm. The symbol $(\star)$ denotes a step which, depending on the properties of the system to control, can be executed in different ways. In the Appendix, we illustrate how to implement the step $(\star)$ for three classes of dynamical systems which encompass a large variety of, yet not all, robotic systems.

\section{Validation and Comparisons}
\label{sec:validation}

This section presents an evaluation and a comparison between different variants of the proposed approach. We start by comparing the partial conformal and the full \ac{qc} mappings, then highlighting the trade-off between computational complexity and robustness of the approach. Then, we will illustrate the difference between mapping the state of the system between polyhedral and ball world, rather than the control input. The discussion in this section will serve as a guide to the choice of algorithm employed in the next section.

\begin{figure}
	\centering
	\includegraphics[width=\linewidth]{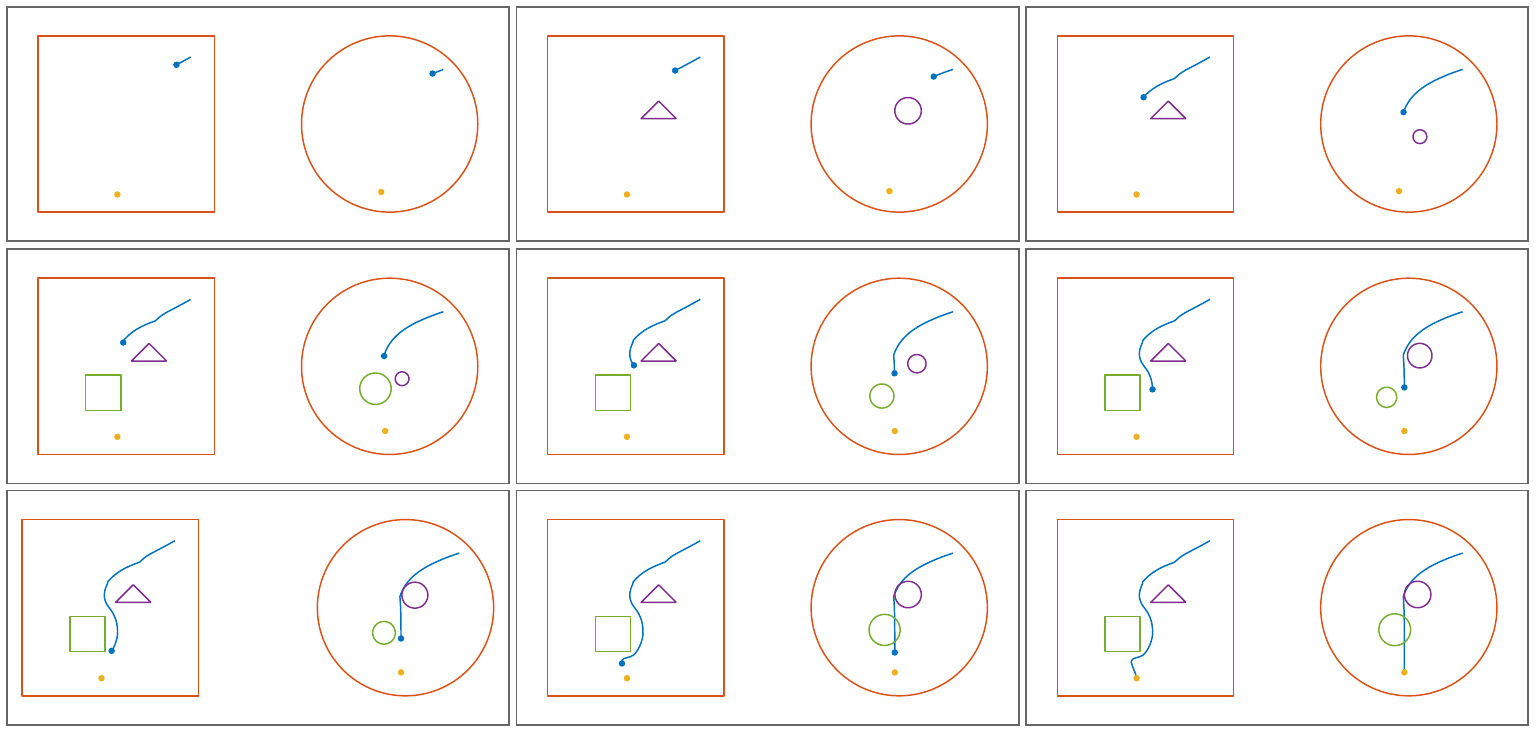}
	\caption{Full \ac{qc} mapping (using the inverse mapping to transform the states). In each panel, the left box is the polyhedral (real) world, and the right circle is the ball world. Notice the motion of the obstacles in the ball world to prevent the state of the robot mapped into the ball world from colliding with them at each point in time.}
	\label{fig:fullQC}
\end{figure}

\subsection{Partial Conformal \textsc{vs} Full Quasi-conformal Mapping}
\label{subsec:fullvspartial}

\begin{figure}
	\centering
	\includegraphics[width=\linewidth]{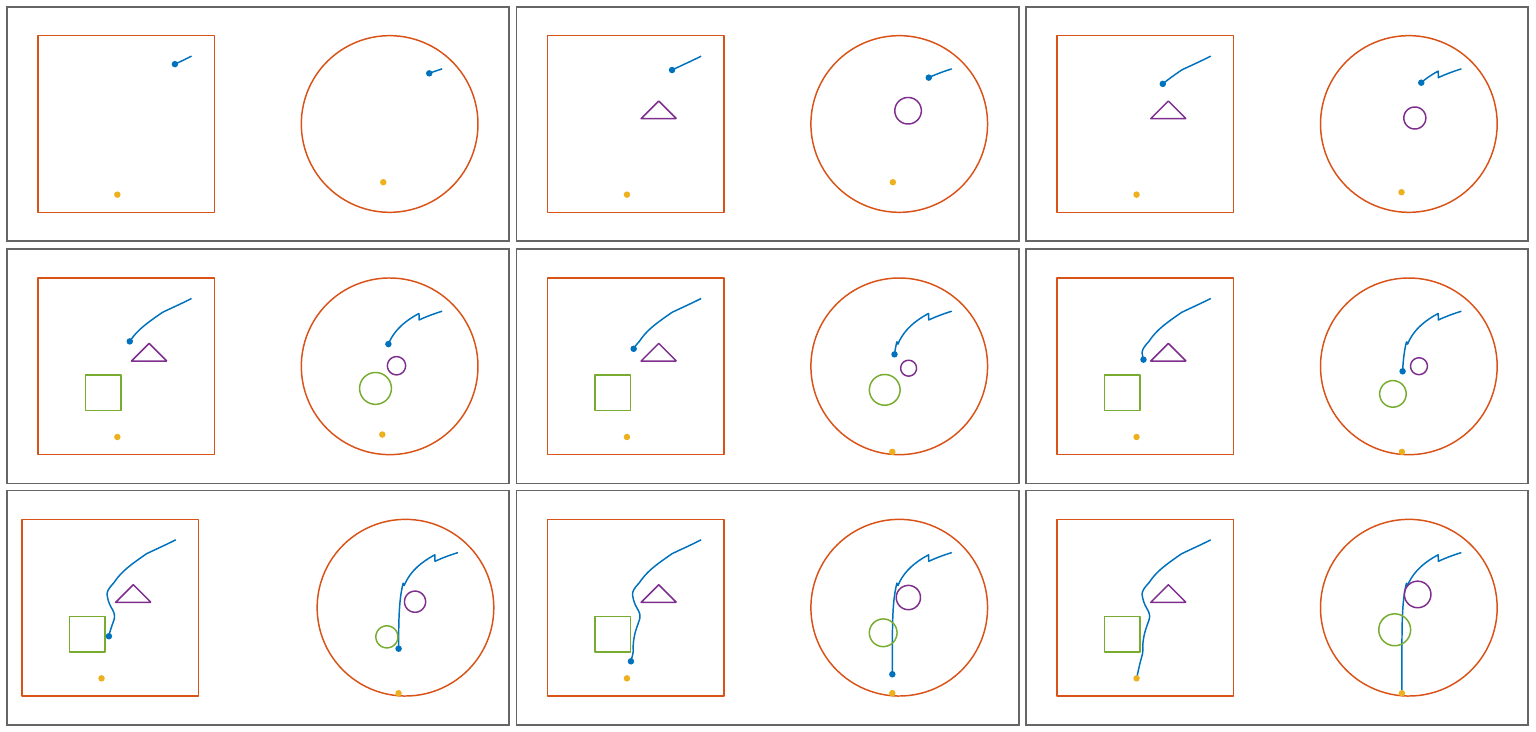}
	\caption{Partial conformal  mapping ($\lambda = 10000$). In each panel, the left box is the polyhedral (real) world, and the right circle is the ball world. Notice the motion of the obstacles in the ball world to prevent the state of the robot mapped into the ball world from colliding with them at each point in time.}
	\label{fig:partialQC}
\end{figure}

\begin{figure}
	\centering
	\includegraphics[width=0.6\columnwidth]{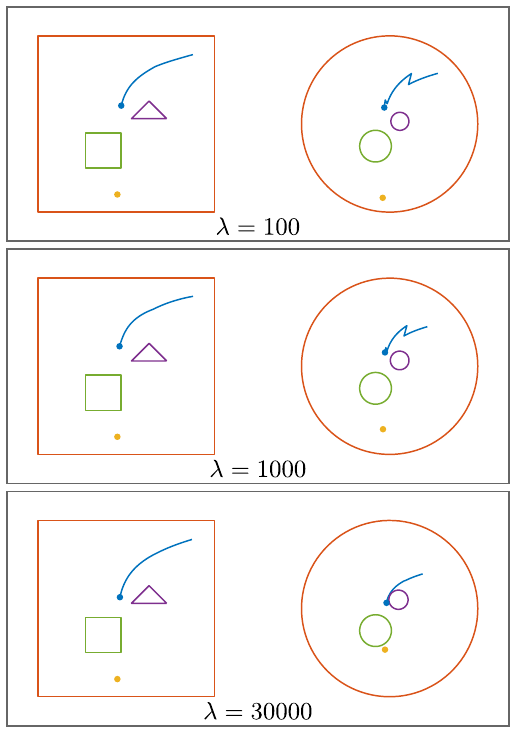}
	\caption{Partial conformal mapping for different values of $\lambda$. In each panel, the left box is the polyhedral (real) world, and the right circle is the ball world.}
	\label{fig:partialQC_lambda}
\end{figure}

We provide a simulation comparison between the Partial Conformal (Sec.~\ref{subsec:partialQC}) and the Full QC (Sec.~\ref{subsec:fullQC}) mappings. To this end, we consider the dynamics
\begin{equation}
	\label{eq:dyn_sys_example}
	\dot{x} = -Ax + u = \begin{bmatrix} 60 & 0 \\ 0 & 10 \end{bmatrix} (x_g - x),
\end{equation}
where the 2-dimensional state vector is defined as $x=[x_1,x_2]\tr$, the goal state is $x_g = -[0.25,2]\tr$, and $u = Ax_g$. We consider a navigation scenario where the state of the dynamics~\eqref{eq:dyn_sys_example} has to reach $x_g$ starting from $x(0) = [2,2]\tr$ while avoiding obstacles that dynamically appear in the state space (the triangle and the square in Fig.~\ref{fig:partialQC}). The navigation task is successfully completed using either the Partial (Fig.~\ref{fig:partialQC}) or the Full (Fig.~\ref{fig:fullQC}) mapping. The Partial is computationally more efficient than the Full mapping, with computational time increasing only when adding new obstacles. However, the Partial mapping is a proper \ac{qc} mapping only for certain values of $\lambda$. Indeed, as shown in Fig.~\ref{fig:partialQC}, the state in the ball world exhibits a discontinuous behavior when new obstacles are added. The jump can be reduced by reducing the effect of the obstacle at larger distances, i.e., by increasing $\lambda$ as shown in Fig.~\ref{fig:partialQC_lambda}, but this requires manual tuning of $\lambda$ at run-time which makes the approach application specific. On the other hand, updating the Full mapping at each iteration is computationally more expensive (Sec.~\ref{subsec:computationvsrobustness}), but it always generates a proper \ac{qc} mapping, and, therefore, smooth state trajectories (Fig.~\ref{fig:fullQC}).

\subsection{Computation \textsc{vs} Robustness Trade-off}
\label{subsec:computationvsrobustness}

\begin{figure}
\centering
\subfigure[][]{\includegraphics[trim={2cm 6.7cm 2.5cm 7cm},clip,width=1\columnwidth]{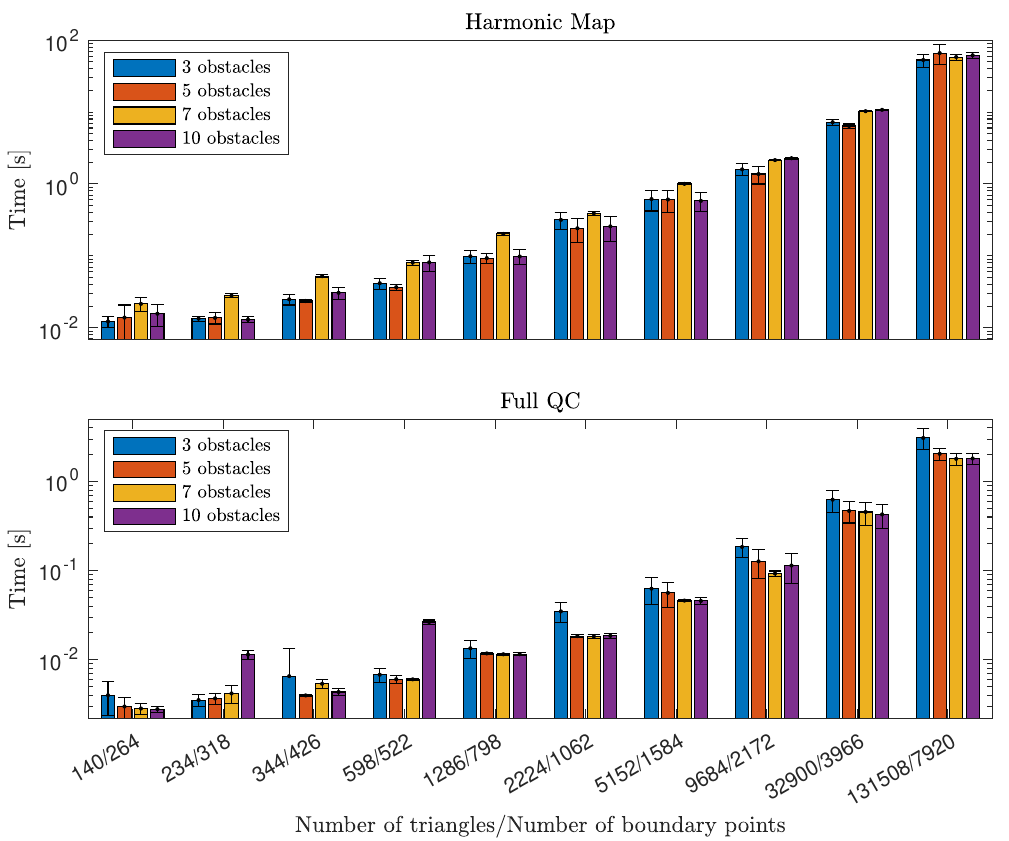}}\\
\subfigure[][]{\includegraphics[trim={2cm 12cm 1.5cm 11.5cm},clip,width=\columnwidth]{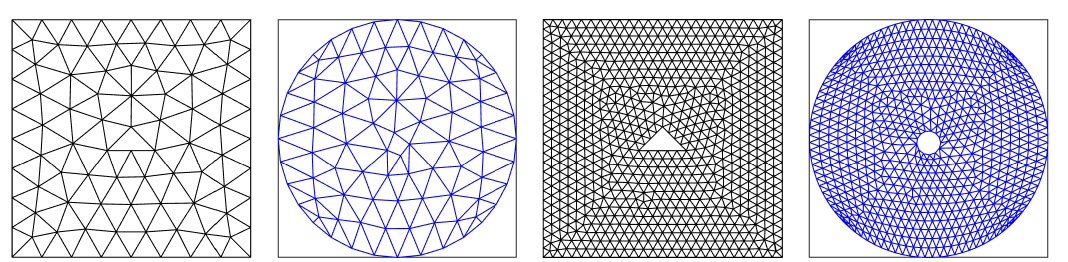}}
\subfigure[][]{\includegraphics[trim={5cm 13.8cm 4.4cm 13.4cm},clip,width=\columnwidth]{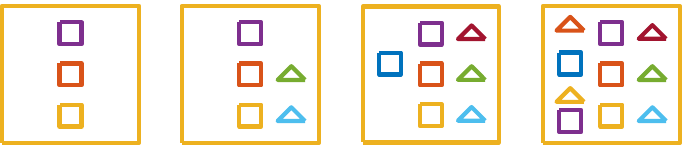}}
\caption{(a) Computational complexity of Harmonic Map~\cite{vlantis2018robot} and the proposed Full QC for different domain approximations. Note the logarithmic scale on the time axis. (b) Mesh domain approximation with $144$ (left) and $10070$ (right) triangles. (c) Polyhedral worlds of increasing complexity considered in this comparison.}
	\label{fig:mapTime}
\end{figure}
In this experiment, we evaluate the computational time of the proposed Full QC mapping, and compare it against the Harmonic Map approach presented in~\cite{vlantis2018robot}. We consider the same box domain (polyhedral world) of Sec.~\ref{subsec:fullvspartial} containing 2 obstacles (a box and a triangle, as shown in Fig.~\ref{fig:mapTime}). Given this domain, we create a triangular mesh using the built-in Matlab\textsuperscript{\textregistered} function \texttt{generateMesh}. The function accepts as input the maximum element size $h_{\mathrm{max}}$ that we sample from the vector $[0.05, 0.1, 0.15, 0.2, 0.3, 0.4, 0.5, 0.6]$. As shown in Fig.~\ref{fig:mapTime}, the value of $h_{\mathrm{max}}$ affects the number of triangles in the mesh, e.g., for  $h_{\mathrm{max}} = 0.6$ the mesh contains $144$ triangles. The Harmonic Map approach approximates the domain and the boundary of each obstacle with a set of points (boundary points). In order to make the two approaches comparable, we compute the number of boundary points for each value of $h_{\mathrm{max}}$ and use it to compute the Harmonic Map. For each value of $h_{\mathrm{max}}$ (or number of boundary points), we compute the Full QC mapping $10$ times and report statistics in Fig.~\ref{fig:mapTime}. Results are obtained using Matlab\textsuperscript{\textregistered} 2018b on a laptop equipped with an Intel i7 $7$\textsuperscript{th} generation CPU and $16\,$GB of DDR4 RAM.

Figure~\ref{fig:mapTime}(a) shows the computation time obtained with Harmonic Map and Full QC approaches in the polyhedral domains shown in Fig.~\ref{fig:mapTime}(c). We observe that the computation time of both Harmonic Map and Full QC is almost independent of the number of obstacles, while it increases with the number of boundary points/triangles used to approximate the domain. For all the considered resolutions, our approach is about one order of magnitude more efficient than the Harmonic Map, which makes it possible to use the Full QC in dynamic scenarios as the one considered in Sec.~\ref{subsec:fullvspartial}. Figure~\ref{fig:mapTime}(b) shows the trade-off between computational complexity (number of triangles) and robustness margin (size of triangles). Fast computation requires fewer triangles. However, having a too coarse mesh may result in a significant loss of representation details. As shown in the left panels of Fig.~\ref{fig:mapTime}(b), with $144$ mesh elements the triangular obstacle is poorly approximated both in the real and in the ball world. A practical solution could be to set the number of triangles (or, equivalently, $h_{\mathrm{max}}$) based on the controller loop time and then set the desired error bounds based on the size of the triangles. This will introduce more error, which can be accounted for in the safety constraints by increasing the safety margin accordingly. Alternatively, one can create a course mesh and refine it locally around the obstacles.

\subsection{State Mapping \textsc{vs} Input Mapping}
\label{subsec:statevsinput}

The safe robot navigation algorithm requires mapping back and forth the state and the control input from the real to the ball world (Sec.~\ref{subsec:control_pol_world}). These mappings can be realized in two ways: \textit{1)} Using the Jacobian of the transformation and its inverse to map control inputs; or \textit{2)} Using the transformation (QC mapping) to map states. 

\begin{figure}
	\centering
	\includegraphics[width=\linewidth]{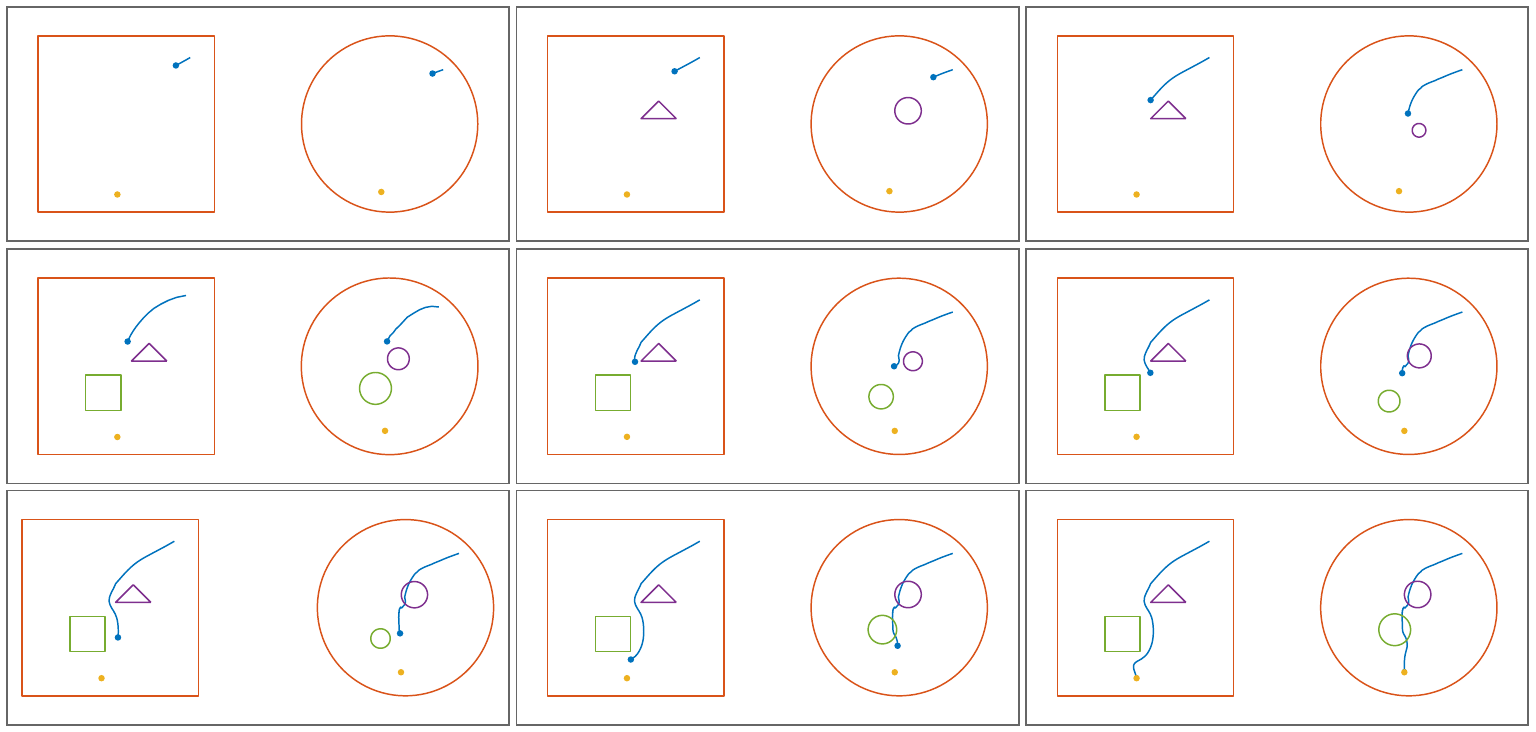}
	\caption{Full \ac{qc} input mapping (using the mapping Jacobian to transform inputs). In each panel, the left box is the polyhedral (real) world, and the right circle is the ball world. As for the previous simulations, notice the motion of the obstacles in the ball world to prevent the state of the robot mapped into the ball world from colliding with them at each point in time.}
	\label{fig:fullQCinput}
\end{figure}

In order to compare the two approaches, we consider the same setup used in Sec.~\ref{subsec:fullvspartial}. Mapping the state of the robot from the ball world to the polyhedral world (Fig.~\ref{fig:fullQC}) is effective only for dynamics with specific properties like differential flatness or feedback linearizability (Sec.~\ref{subsec:control_pol_world}). Instead, mapping control inputs using the Jacobian and its inverse is a more general approach that does not require further assumptions on the dynamics beyond smoothness. Results obtained with this approach are shown in Fig.~\ref{fig:fullQCinput}. In the considered scenario, both approaches are able to successfully complete the navigation task.

When it comes to numerical implementation, the two approaches exhibit different behaviors. In particular, the approach proposed to compute the \ac{qc} mapping ensures a certain level of smoothness. However, the mapping, especially close to the obstacle boundary, can be significantly deformed. Since the Jacobian is numerically computed using finite differences, this may result in a close to singular Jacobian matrix, and, as a consequence, in an almost unbounded inverse mapping. To prevent these issues, we used a varying step numerical approach to compute derivatives, but this comes at an extra computational cost. On the other hand, mapping the state only requires the mapping and its inverse that are smooth everywhere in the state space.

\subsection{Cluttered Office Workspace}
\begin{figure}
	\centering
\includegraphics[trim={6cm 9cm 5cm 8.5cm},clip,width=0.8\columnwidth]{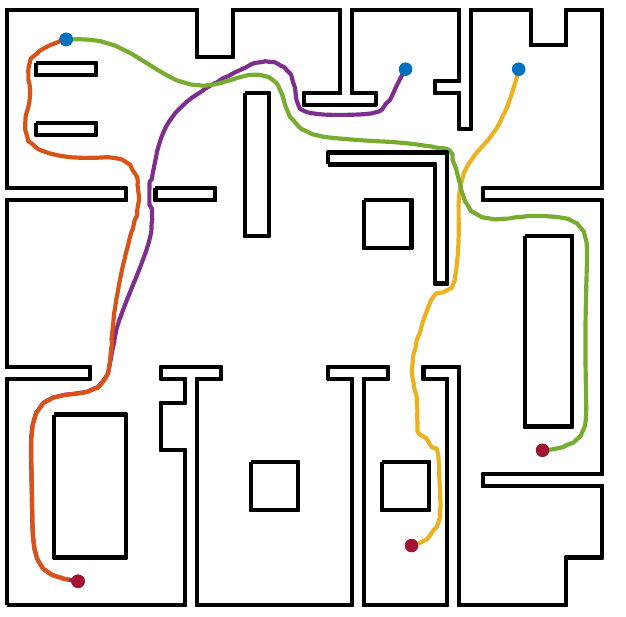}
	\caption{Results obtained with the Full QC mapping in a cluttered office workspace. Blue/red bullets indicate start/goal positions.}
	\label{fig:officeNavigation}
\end{figure}
In this experiment, we demonstrate that the Full QC approach can be used to navigate cluttered domains with a complex boundary. To this end, we construct a cluttered office workspace (Fig.~\ref{fig:officeNavigation}) similar to the one used in~\cite{vlantis2018robot,fan2022robot}. This domain is designed to reproduce a realistic office workspace, and it is complex to navigate as it contains $10$ objects of different shape and size and it has complex boundaries. As shown in Fig.~\ref{fig:officeNavigation}, the Full QC approach is able to navigate the office from different initial to different final positions.

\section{Robotics Applications}
\label{sec:robotic_applications}
This section presents experimental results in typical robotics applications, both in simulation and on real devices.
\subsection{Avoiding low manipulability areas}
\begin{figure}
	\centering
	\includegraphics[width=\columnwidth,trim={1cm 0.1cm 1cm 0.5cm},clip]{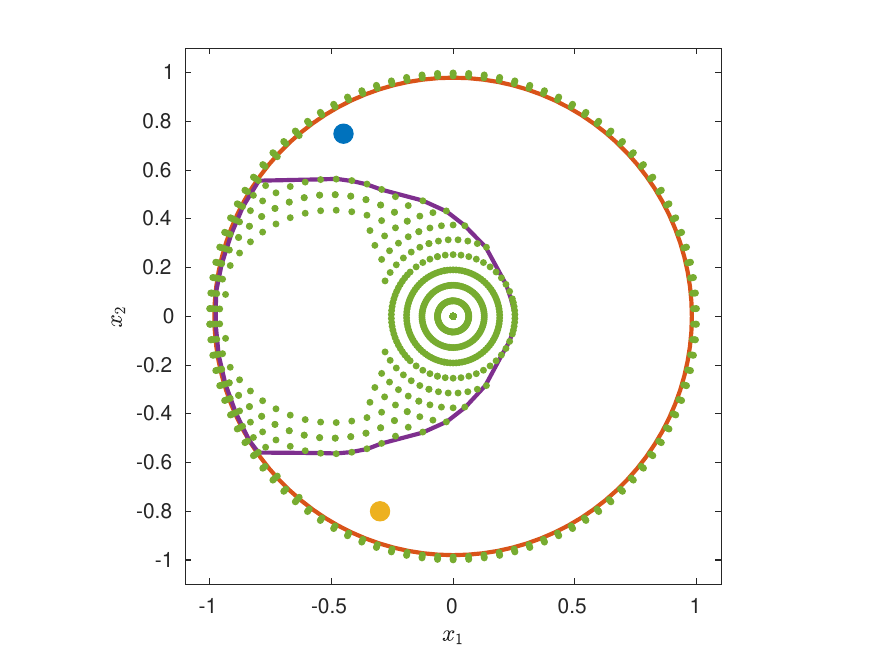}
	\caption{The workspace of an RR planar manipulator. The blue dot is the starting point, the yellow dot is the goal, while the green dots are regions of low manipulability ($\mu < 0.1$).}
	\label{fig:2R_Workspace}
\end{figure}

In this simulation, we consider a planar two-link manipulator that has to move its end-effector between two points (blue and yellow dots in Fig.~\ref{fig:2R_Workspace}) while avoiding a region of low manipulability (the region within the purple boundary). The manipulator has two links of equal length, i.e., $l_1 = l_2 = 0.5\,$m. The joint ranges are limited to $q_1, q_2 \in [\underline{q}, \overline{q}] = [-\pi,\pi]\,$rad. Given a joint configuration $q=[q_1,q_2]^\top$, and the corresponding manipulator Jacobian $J(q)$, the manipulability is computed as
\begin{equation}
    \label{eq:manipulability}
    \mu = k_{\mu} \sqrt{\det(J(q)J^\top(q))},
\end{equation}
where the gain $k_{\mu}$ is used to reduce the manipulability while approaching the joint limits, and it is computed as
\begin{equation}
    \label{eq:manipulability_gain}
    k_{\mu} = 1 - \exp{\left(-100\prod_{i=1}^2 \dfrac{(q_i - \underline{q})(\overline{q} - q_i)}{(\overline{q} - \underline{q})^2}\right)}.
\end{equation}

\begin{figure}
	\centering
    \includegraphics[width=\linewidth]{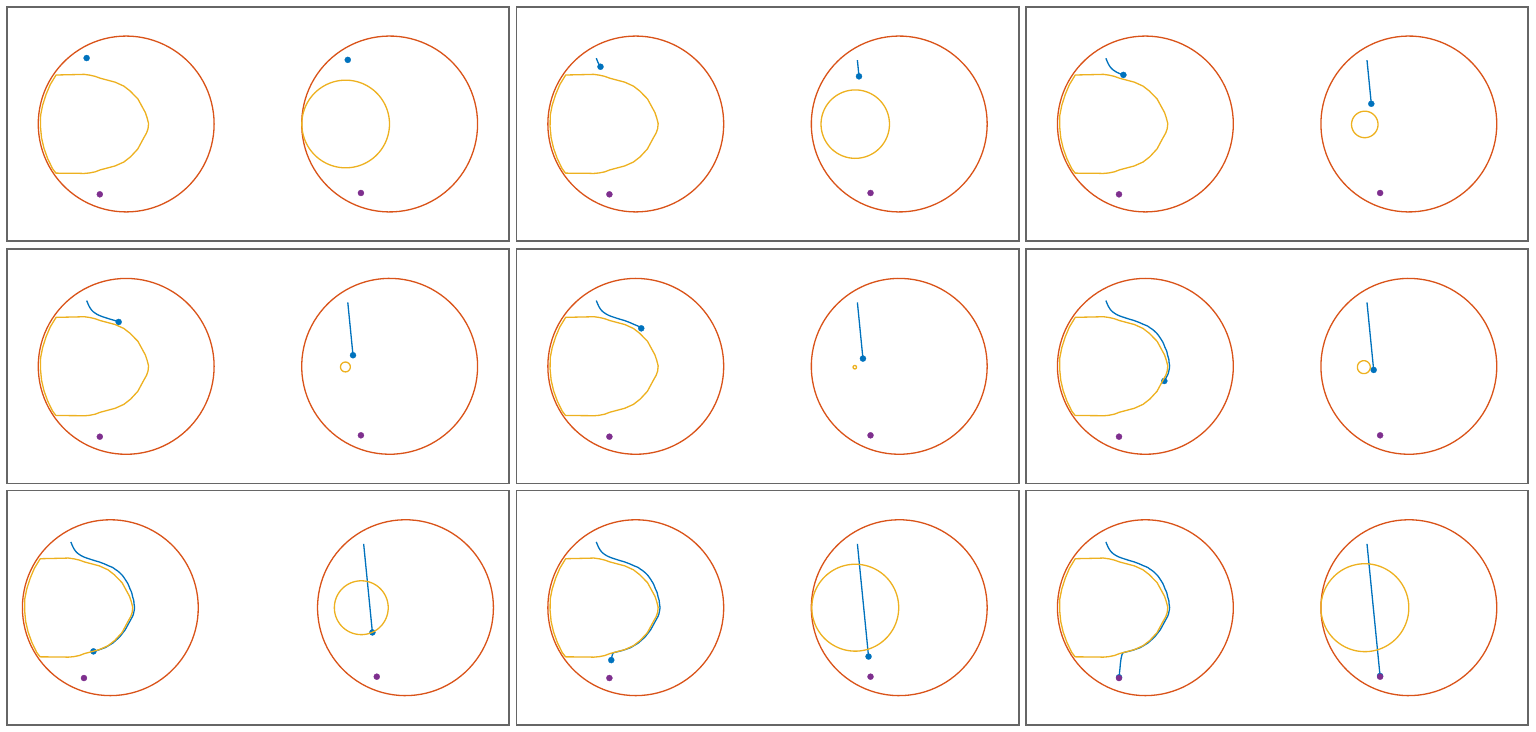}
    \caption{Avoiding low manipulability areas with the full \ac{qc} mapping. In each panel, the left circle is the polyhedral (real) world, representing the robot's workspace, and the right circle is the ball world. Notice the motion of the obstacles in the ball world to prevent the state of the robot mapped into the ball world from colliding with them at each point in time.}
	\label{fig:2R_snapshots}
\end{figure}

The region of low manipulability (green dots in Fig.~\ref{fig:2R_Workspace}) is computed by uniformly sampling ($50\times 50$ grid) the configuration space of the robot, computing the manipulability $\mu$ at each point using~\eqref{eq:manipulability}, and marking the point as low manipulability if $\mu < 0.1$. Then, we create the boundary of the polyhedral world (orange circle in Fig.~\ref{fig:2R_Workspace}) as a circle of radius $0.98\,$m. This is enough to exclude low manipulability configurations corresponding to the manipulator fully stretched. Other low manipulability areas\footnote{This boundary is computed using the built-in Matlab\textsuperscript{\textregistered} function \texttt{boundary}.} (purple boundary in Fig.~\ref{fig:2R_Workspace}) are considered as an obstacle to be avoided. As shown in Fig.~\ref{fig:2R_snapshots}, using the Full QC mapping and only reducing the obstacle's radius in the ball world, the safe reactive robot navigation algorithm successfully executes the task (reach a goal joint configuration) while preserving safety (avoid low manipulability regions). 

\subsection{Avoiding forbidden areas with a pan--tilt camera}
\begin{figure}
	\centering
    \includegraphics[width=\linewidth]{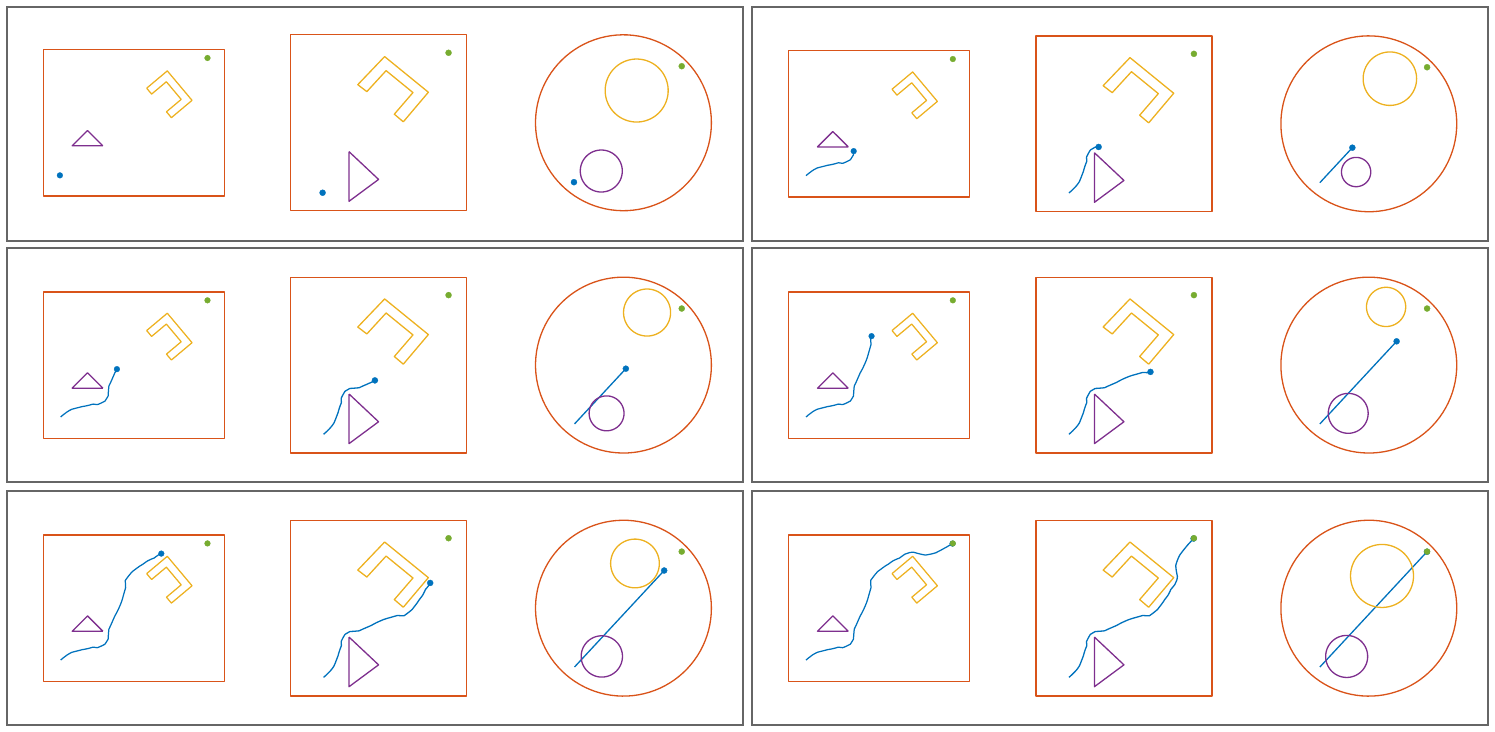}
    \caption{Avoiding forbidden areas with a pan-tilt camera performing a surveillance task. In each panel, the left rectangle is the $xy$ plane, the middle square is the pan-tilt angles (the polyhedral world), and the circle is the ball world. Also in this figure, the motion of the obstacles in the ball world is what prevents the state of the robot mapped into the ball world from colliding with them at each point in time.}
	\label{fig:PT_snapshots}
\end{figure}

We consider a surveillance task where a pan--tilt camera has to scan the space between two configurations (blue and green dots in Fig.~\ref{fig:PT_snapshots}) while avoiding two forbidden areas (purple triangle and yellow polygon in Fig.~\ref{fig:PT_snapshots}). In practice, inspecting inside the forbidden areas may correspond to privacy violating areas, e.g., if the area contains the windows of a private building.

The considered pan--tilt camera has $3$ links, namely $l_0 = 0\,$m and $l_1 = l_2 = 1\,$m. Also in this case, the joint ranges are limited to $q_1, q_2 \in [\underline{q}, \overline{q}] = [-\pi,\pi]\,$rad.  Given the pan--tilt angles $q = [q_1, q_2]^{\top}$, the 3D position of the camera (end-effector) is computed as
\begin{equation}
   p = \begin{bmatrix}
        x \\
        y \\
        z 
    \end{bmatrix} =
    \begin{bmatrix}
        l_1 + l_2\sin(q_2) \\
        l_2\sin(q_1)\cos(q_2) \\
        l_0 - l_2\cos(q_1)\cos(q_2) 
    \end{bmatrix}.
    \label{eq:fk_pan_tilt}
\end{equation}
The inverse mapping from position $p$ to joint angles $q$ can be computed as
\begin{equation}
   q = \begin{bmatrix}
        q_1 \\
        q_2
    \end{bmatrix} =
    \begin{bmatrix}
        \mathrm{atan2}(y, l_0-z)  \\
        \mathrm{atan2}\left(\dfrac{x-l_1}{l_2}, \sqrt{ \dfrac{(y^2 + (l_0-z)^2)}{l_2^2}}\right) 
    \end{bmatrix} .
    \label{eq:ik_pan_tilt}
\end{equation}

The forbidden areas in Fig.~\ref{fig:PT_snapshots} are defined in task space. In order to obtain the polyhedral world in Fig.~\ref{fig:PT_snapshots}, we map the forbidden areas to the configuration space using~\eqref{eq:ik_pan_tilt}. The polyhedral world is then mapped into the ball world using the Full QC mapping. Results in Fig.~\ref{fig:PT_snapshots} show that the proposed safe robot navigation algorithm successfully executes the task (reach a goal Cartesian position) while preserving safety (avoid forbidden areas). The experiment also shows that the proposed approach can be combined with other smooth mappings (in this case the inverse kinematics in~\eqref{eq:ik_pan_tilt})  to solve specific problems.

\begin{figure*}
\centering
\subfigure[][Results obtained without mapping the free space in the polyhedral world to the free space in a ball world.  The controller preserves only the safety.]{%
\includegraphics[width=0.16\textwidth,trim={12cm 2cm 10cm 3.5cm},clip]{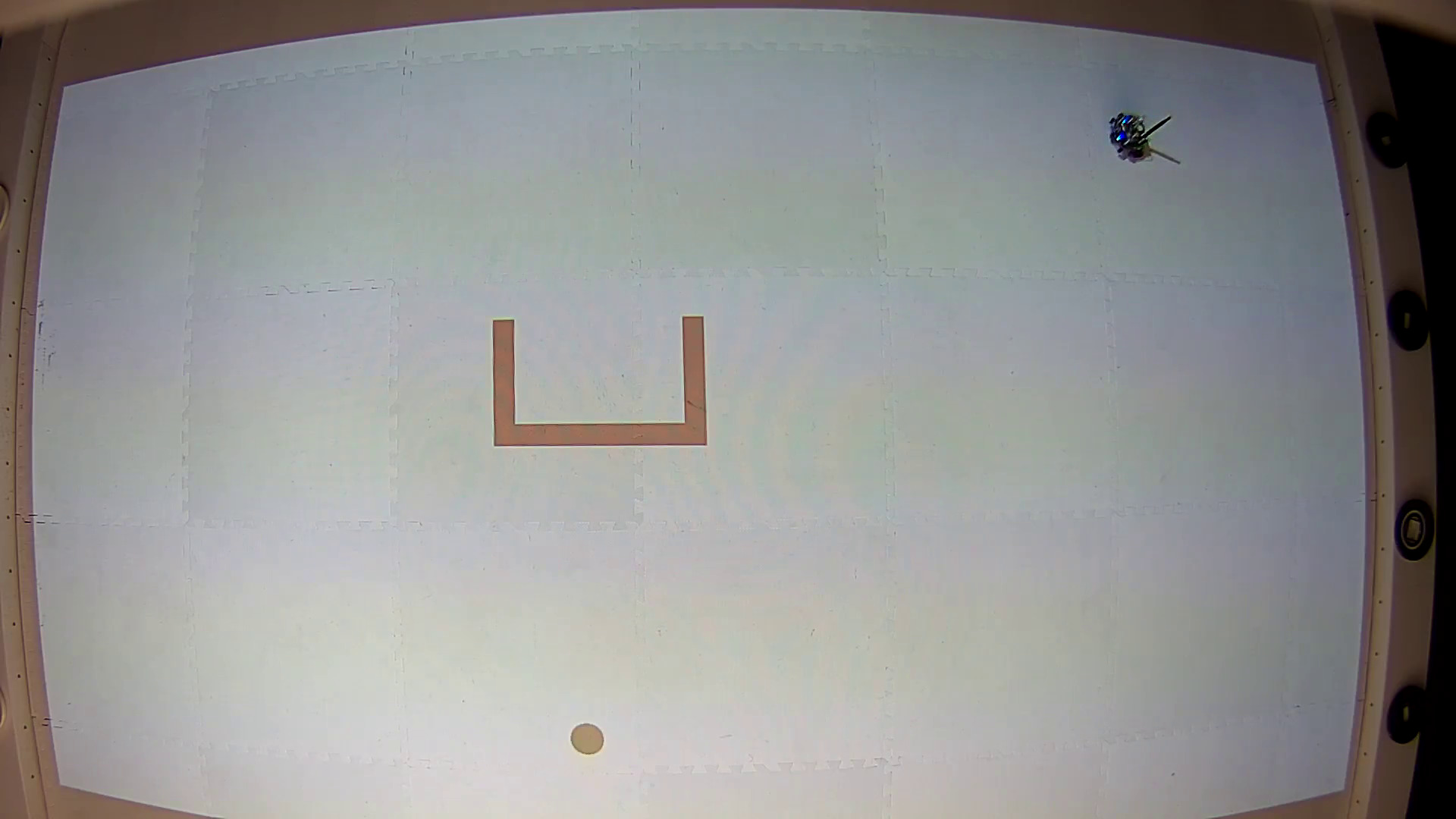}%
\includegraphics[width=0.16\textwidth,trim={12cm 2cm 10cm 3.5cm},clip]{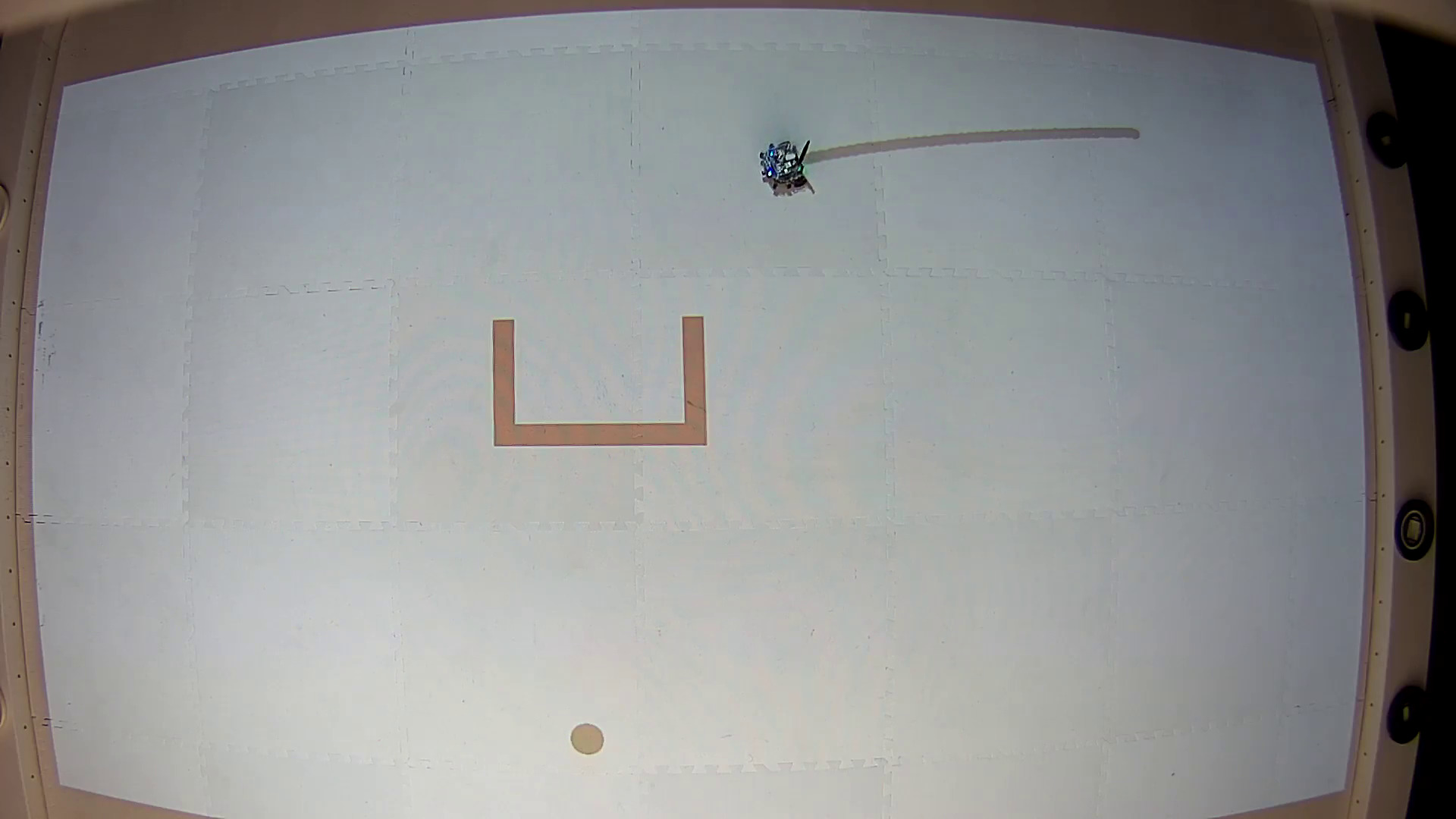}%
\includegraphics[width=0.16\textwidth,trim={12cm 2cm 10cm 3.5cm},clip]{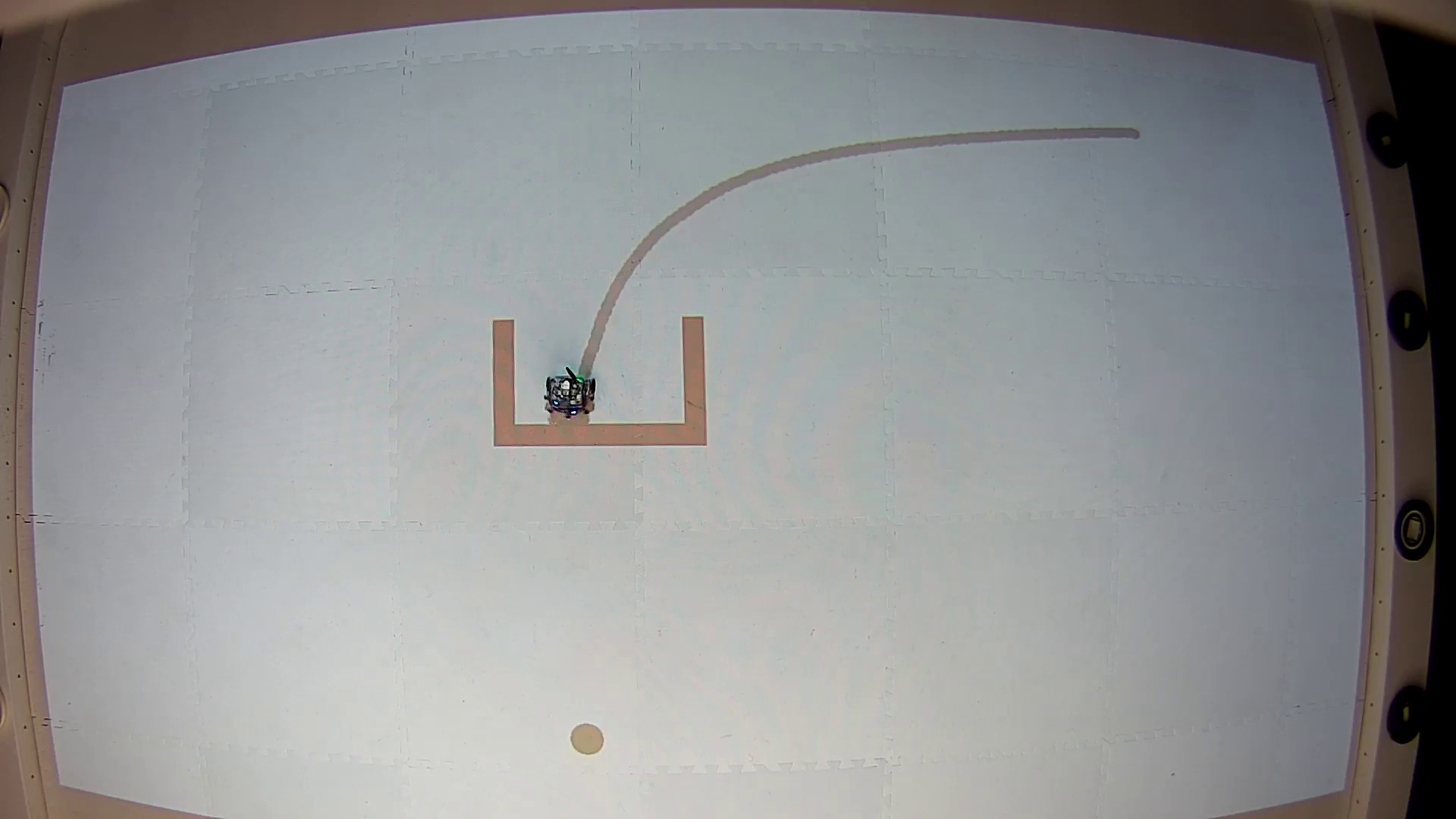}}%
\hfill%
\subfigure[][Results obtained using the partial conformal mapping. The controller preserves both safety and stability.]{%
\includegraphics[width=0.16\textwidth,trim={12cm 2cm 10cm 3.5cm},clip]{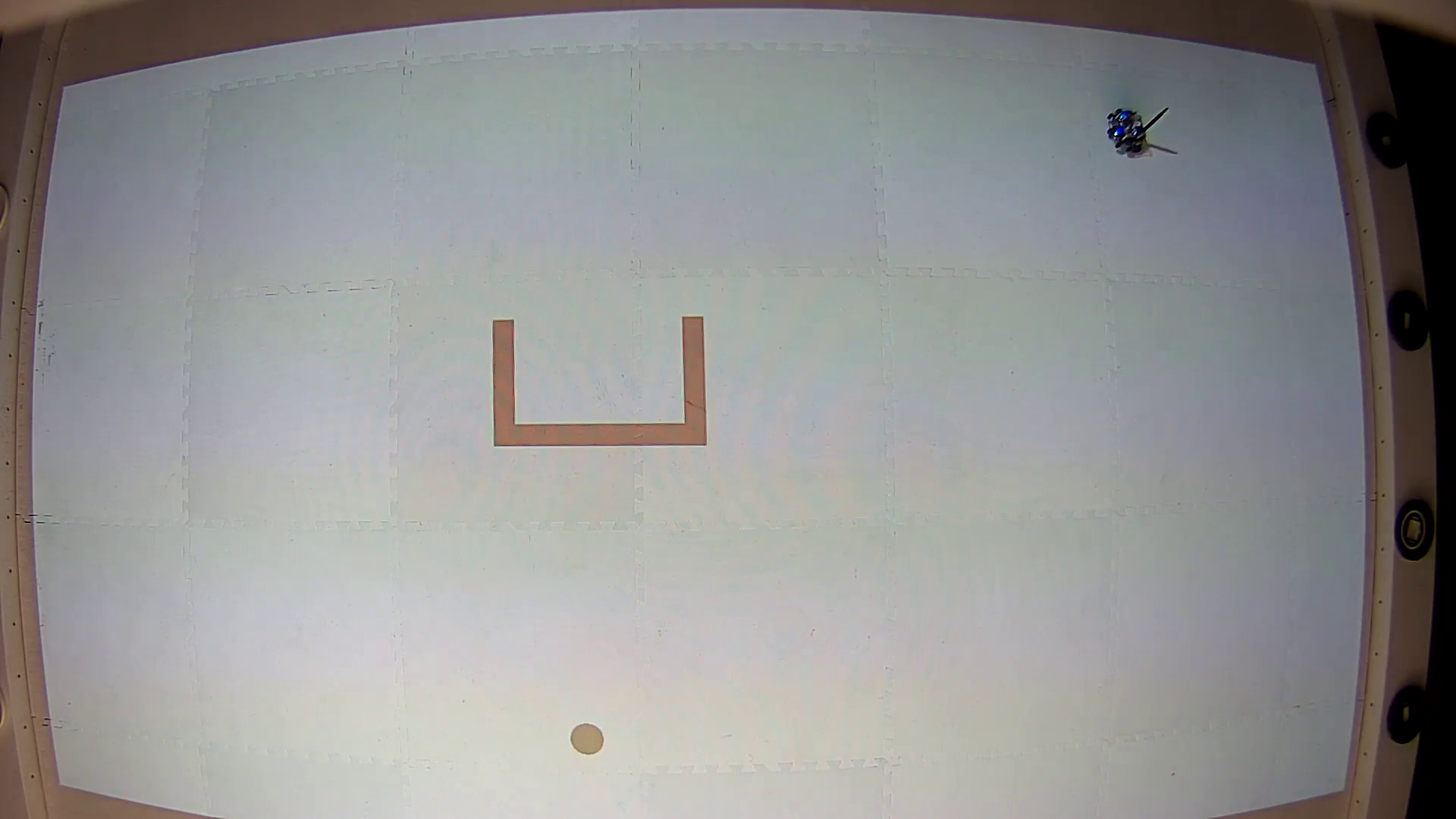}%
\includegraphics[width=0.16\textwidth,trim={12cm 2cm 10cm 3.5cm},clip]{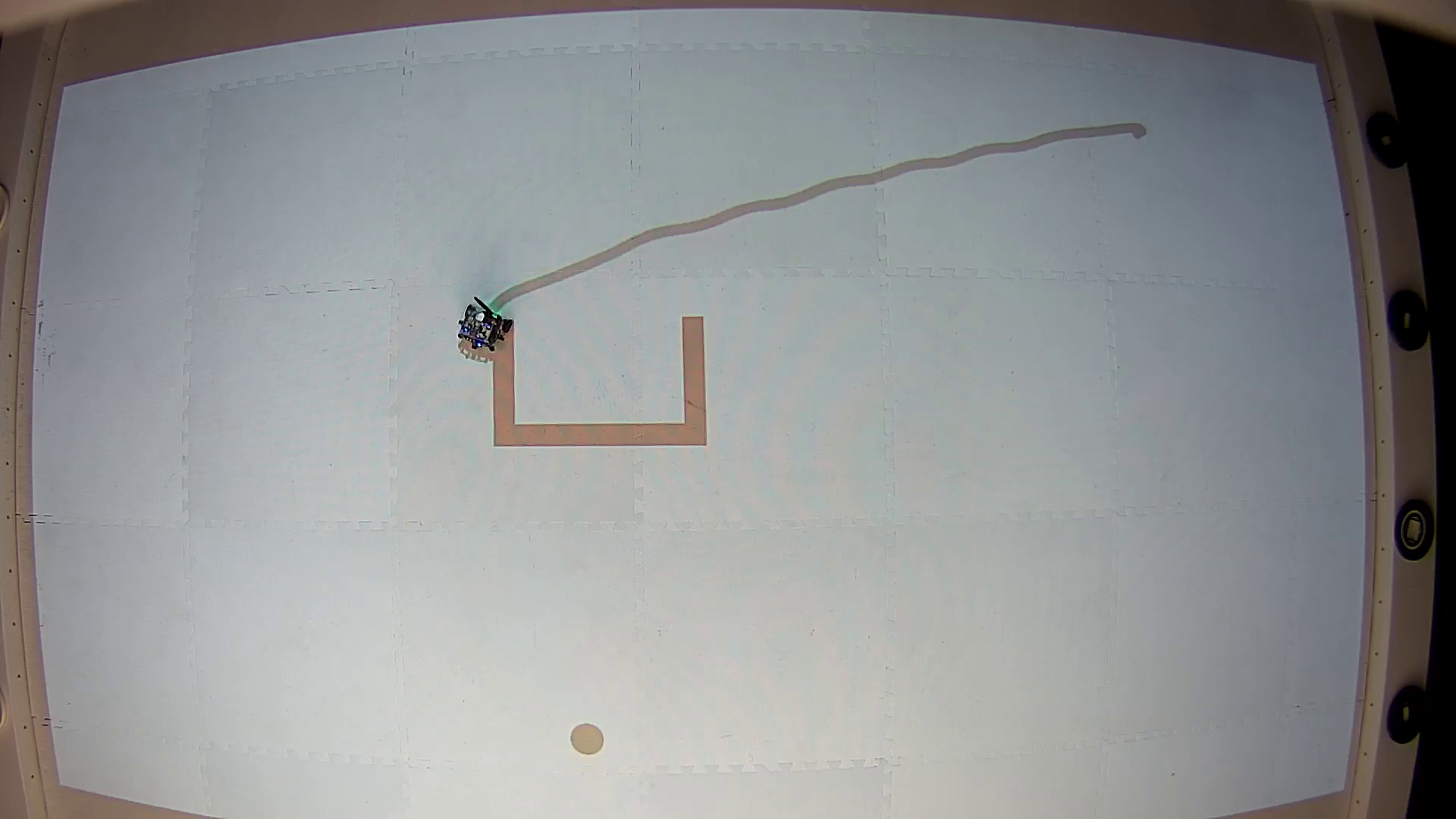}%
\includegraphics[width=0.16\textwidth,trim={12cm 2cm 10cm 3.5cm},clip]{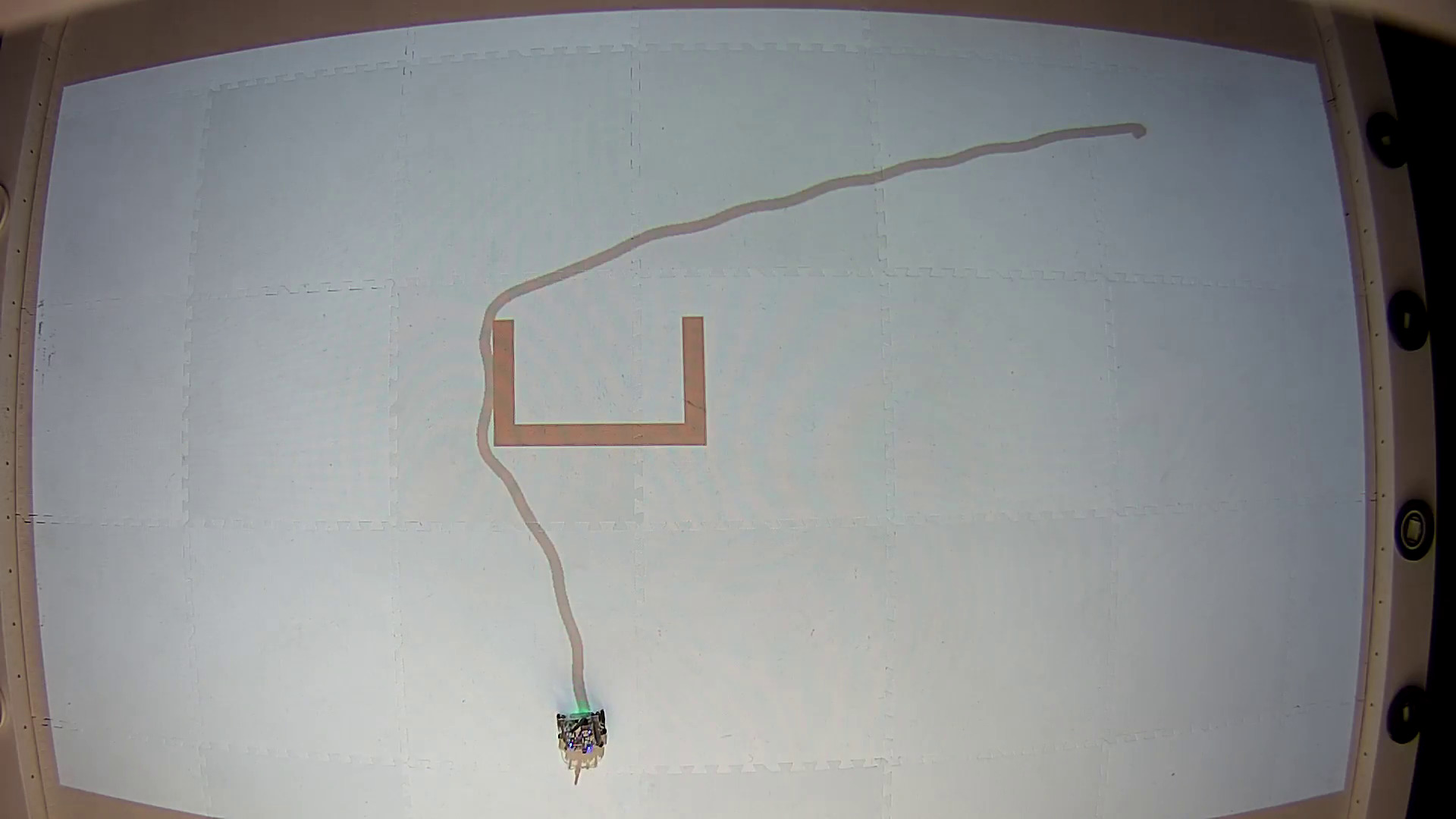}}
\caption{Snapshots of the robotarium experiments.}
\label{fig:robotarium}
\end{figure*}

\subsection{Navigation in the Robotarium}
\label{subsec:robotarium}
In this experiment, we consider a mobile robot that has to navigate between two points while avoiding a concave obstacle (Fig.~\ref{fig:robotarium}). Experiments are performed inside the Robotarium~\cite{wilson2020robotarium}. Using
the traditional \ac{cbf}-\ac{qp} formulation directly in the polyhedral world generates an undesired equilibrium
point (Fig.~\ref{fig:robotarium}(a)), which is eliminated using the proposed approach (Fig.~\ref{fig:robotarium}(b)).

\section{Conclusions}\label{sec:conclusions}
In this paper, we presented a computationally efficient approach for safe reactive robot navigation that exploits quasi-conformal mappings and control barrier functions. Quasi-conformal mappings are used to compute a smooth mapping between the state space of the robot, containing possibly multiple and non-convex unsafe regions, into a space where the unsafe regions are either closed balls or the complement of open balls. Quasi-conformal mappings allow us to have mild assumptions on the shape of the unsafe regions and, importantly, do not require an analytical representation of the unsafe regions. We presented and analyzed two versions of the proposed mapping, a partial and a full one. The full mapping updates the entire workspace at each iteration of the control loop, while the partial mapping computes the transformation for each unsafe region independently and then combines them. The full mapping is computationally more intensive, but the resulting mapping is guaranteed to be a diffeomorphism. On the contrary, the partial mapping is more efficient---the mapping can be locally updated as novel unsafe regions appear or disappear, but it requires careful tuning of a parameter to ensure a proper diffeomorphic mapping. Given the mapping, we design a controller in a ball world---a transformed robot workspace where all obstacles are closed balls---that shifts and shrinks unsafe ball-shaped regions to ensure safe navigation, relying on the control barrier function formalism. The controller and the two mapping strategies have been validated through simulated and real experiments, showing promising results. 

Our future research will focus on improving the efficiency of the full quasi-conformal mapping. A possibility is to consider meshes of different granularity, i.e., coarse in the free space and fine close to unsafe regions. We also plan to test our algorithm in the presence of partial occlusions, moving obstacles, and dynamic models of robotic systems. Finally, although our proposed algorithm has been shown to be applicable to a large variety of dynamical system models, we are interested in extending it to design a computationally efficient safety layer for existing tracking controllers.

\appendix

\label{subsec:control_pol_world}
This Appendix shows how the reactive navigation strategy developed in this paper can be applied to a large variety of robotic systems.
\subsection{Dynamics (Near-identity) Diffeomorphic to a Single Integrator}

For systems that are diffeomorphic (resp. near-identity diffeomorphic) to a single integrator \cite{olfati2002near}, one can apply a (nonlinear) change of coordinates to the state such the transformed state (resp. part of the state) has linear dynamics. As an example of a system near-identity diffeomorphic to a single integrator, consider the following unicycle dynamics:
\begin{equation}
	\begin{cases}
		\dot p_x = v \cos\theta,\\
		\dot p_y = v \sin\theta,\\
		\dot \theta = \omega,
	\end{cases}
	\label{eq:unicycle}
\end{equation}
where $p_x,p_y$ denote the position of the unicycle, $\theta$ its orientation, and $v,\omega$ are the longitudinal and angular velocity inputs. Using the change of coordinates
\[
z = \begin{bmatrix}
	p_x\\
	p_y
\end{bmatrix} + l \begin{bmatrix}
	\cos\theta\\
	\sin\theta
\end{bmatrix},
\]
leads to
\begin{equation}
	\dot z = \begin{bmatrix}
		\cos\theta & -l \sin\theta\\
		\sin\theta & l \cos\theta
	\end{bmatrix} \begin{bmatrix}
		v\\
		\omega
	\end{bmatrix}.
	\label{eq:unicycleinputmapping}
\end{equation}
The dynamics of $z$ may be controlled using single integrator dynamics to achieve the desired $\dot x^{(k)}$ in Step 10 of Algorithm~\ref{alg:safemultipleconcave}, from which the $v,\omega$ can be computed using \eqref{eq:unicycleinputmapping} in order to control the unicycle dynamics. This approach was taken in Section~\ref{subsec:robotarium} to implement the safe robot navigation algorithm using a differential drive robot.

\subsection{Feedback Linearizable Dynamics}

For feedback linearizable system dynamics \cite{khalil2015nonlinear}, one can apply a (nonlinear) change of coordinates $y=T(x)$ such that there exists a (nonlinear) feedback control law that cancels the nonlinearities of the system. The feedback linearized dynamics are represented by a chain of integrators. As far as the application of the approach presented in this paper is concerned, this class of systems represents a generalization of systems (near-identity) diffeomorphic to a single integrator. Step 11 in Algorithm~\ref{alg:safemultipleconcave} can be executed by designing a stabilizing state-feedback controller with a feedforward term equal to the desired $\dot x^{(k)}$, given in Step 10 of Algorithm~\ref{alg:safemultipleconcave}.

\subsection{Differentially Flat Dynamics}

For differentially flat systems, there exists an output thereof---the flat output---such that their state and input can be determined from the output without integration \cite{fliess1995flatness}. This way, we can proceed similarly to the case of feedback linearizable dynamics and design a stabilizing flat-output-feedback controller with a feedforward term equal to the desired $\dot x^{(k)}$, given in Step 10 of Algorithm~\ref{alg:safemultipleconcave}.

\bibliographystyle{IEEEtran}
\bibliography{bib/references}

\end{document}

%% file: acronyms.tex
\newacro{qp}[QP]{Quadratic Program}
\newacro{qc}[QC]{Quasi-Conformal}
\newacro{cbf}[CBF]{Control Barrier Function}
\newacro{zbf}[ZBF]{Zeroing Barrier Function}
\newacro{zcbf}[ZCBF]{Zeroing Control Barrier Function}
\newacro{ds}[DS]{Dynamical System}

%% file: main.bbl
\begin{thebibliography}{10}
\providecommand{\url}[1]{#1}
\csname url@samestyle\endcsname
\providecommand{\newblock}{\relax}
\providecommand{\bibinfo}[2]{#2}
\providecommand{\BIBentrySTDinterwordspacing}{\spaceskip=0pt\relax}
\providecommand{\BIBentryALTinterwordstretchfactor}{4}
\providecommand{\BIBentryALTinterwordspacing}{\spaceskip=\fontdimen2\font plus
\BIBentryALTinterwordstretchfactor\fontdimen3\font minus
  \fontdimen4\font\relax}
\providecommand{\BIBforeignlanguage}[2]{{%
\expandafter\ifx\csname l@#1\endcsname\relax
\typeout{** WARNING: IEEEtran.bst: No hyphenation pattern has been}%
\typeout{** loaded for the language `#1'. Using the pattern for}%
\typeout{** the default language instead.}%
\else
\language=\csname l@#1\endcsname
\fi
#2}}
\providecommand{\BIBdecl}{\relax}
\BIBdecl

\bibitem{wang2017safety}
L.~Wang, A.~D. Ames, and M.~Egerstedt, ``Safety barrier certificates for
  collisions-free multirobot systems,'' \emph{IEEE Trans. Robot.}, vol.~33,
  no.~3, pp. 661--674, 2017.

\bibitem{notomista2020persistification}
G.~{Notomista} and M.~{Egerstedt}, ``Persistification of robotic tasks,''
  \emph{IEEE Trans. Control Syst. Technol.}, vol.~29, no.~2, pp. 756--767,
  2020.

\bibitem{aswani2013provably}
A.~Aswani, H.~Gonzalez, S.~S. Sastry, and C.~Tomlin, ``Provably safe and robust
  learning-based model predictive control,'' \emph{Automatica}, vol.~49, no.~5,
  pp. 1216--1226, 2013.

\bibitem{saveriano2019learning}
M.~{Saveriano} and D.~{Lee}, ``Learning barrier functions for constrained
  motion planning with dynamical systems,'' in \emph{IEEE/RSJ International
  Conference on Intelligent Robots and Systems}, 2019, pp. 112--119.

\bibitem{petti2005safe}
S.~Petti and T.~Fraichard, ``Safe motion planning in dynamic environments,'' in
  \emph{IEEE/RSJ International Conference on Intelligent Robots and Systems},
  2005, pp. 2210--2215.

\bibitem{loizou2017navigation}
S.~G. Loizou, ``The navigation transformation,'' \emph{IEEE Trans. Robot.},
  vol.~33, no.~6, pp. 1516--1523, 2017.

\bibitem{thyri2020reactive}
E.~H. Thyri, E.~A. Basso, M.~Breivik, K.~Y. Pettersen, R.~Skjetne, and A.~M.
  Lekkas, ``Reactive collision avoidance for asvs based on control barrier
  functions,'' in \emph{IEEE Conference on Control Technology and
  Applications}, 2020, pp. 380--387.

\bibitem{jha2018safe}
S.~Jha, V.~Raman, D.~Sadigh, and S.~A. Seshia, ``Safe autonomy under perception
  uncertainty using chance-constrained temporal logic,'' \emph{J. Autom.
  Reason.}, vol.~60, no.~1, pp. 43--62, 2018.

\bibitem{notomista2020long}
G.~Notomista, ``Long-duration robot autonomy: From control algorithms to robot
  design,'' Ph.D. dissertation, Georgia Institute of Technology, 2020.

\bibitem{cheng2019end}
R.~Cheng, G.~Orosz, R.~M. Murray, and J.~W. Burdick, ``End-to-end safe
  reinforcement learning through barrier functions for safety-critical
  continuous control tasks,'' in \emph{AAAI Conference on Artificial
  Intelligence}, vol.~33, no.~1, 2019, pp. 3387--3395.

\bibitem{emam2022safe}
Y.~Emam, G.~Notomista, P.~Glotfelter, Z.~Kira, and M.~Egerstedt, ``Safe
  reinforcement learning using robust control barrier functions,'' \emph{IEEE
  Robot. Autom. Lett.}, no.~99, pp. 1--8, 2022.

\bibitem{ohnishi2019barrier}
M.~Ohnishi, L.~Wang, G.~Notomista, and M.~Egerstedt, ``Barrier-certified
  adaptive reinforcement learning with applications to brushbot navigation,''
  \emph{IEEE Transactions on robotics}, vol.~35, no.~5, pp. 1186--1205, 2019.

\bibitem{notomista2023constrained}
G.~Notomista, ``A constrained-optimization approach to the execution of
  prioritized stacks of learned multi-robot tasks,'' \emph{arXiv preprint
  arXiv:2301.05346}, 2023.

\bibitem{hsu2023safety}
K.-C. Hsu, H.~Hu, and J.~F. Fisac, ``The safety filter: A unified view of
  safety-critical control in autonomous systems,'' \emph{Annual Review of
  Control, Robotics, and Autonomous Systems}, vol.~7, 2023.

\bibitem{ames2019control}
A.~D. Ames, S.~Coogan, M.~Egerstedt, G.~Notomista, K.~Sreenath, and P.~Tabuada,
  ``Control barrier functions: Theory and applications,'' in \emph{2019 18th
  European Control Conference (ECC)}.\hskip 1em plus 0.5em minus 0.4em\relax
  IEEE, 2019, pp. 3420--3431.

\bibitem{breeden2022predictive}
J.~Breeden and D.~Panagou, ``Predictive control barrier functions for online
  safety critical control,'' in \emph{2022 IEEE 61st Conference on Decision and
  Control (CDC)}.\hskip 1em plus 0.5em minus 0.4em\relax IEEE, 2022, pp.
  924--931.

\bibitem{reis2020control}
M.~F. {Reis}, A.~P. {Aguiar}, and P.~{Tabuada}, ``Control barrier
  function-based quadratic programs introduce undesirable asymptotically stable
  equilibria,'' \emph{IEEE Control Systems Letters}, vol.~5, no.~2, pp.
  731--736, 2021.

\bibitem{rimon1990exact}
E.~Rimon and D.~E. Koditschek, ``Exact robot navigation in geometrically
  complicated but topologically simple spaces,'' in \emph{Proceedings., IEEE
  International Conference on Robotics and Automation}.\hskip 1em plus 0.5em
  minus 0.4em\relax IEEE, 1990, pp. 1937--1942.

\bibitem{rimon1991construction}
------, ``The construction of analytic diffeomorphisms for exact robot
  navigation on star worlds,'' \emph{Trans. Am. Math. Soc.}, vol. 327, no.~1,
  pp. 71--116, 1991.

\bibitem{rimon1992exact}
------, ``Exact robot navigation using artificial potential functions,''
  \emph{IEEE Trans. Robot. Autom.}, vol.~8, no.~5, pp. 501--518, 1992.

\bibitem{belta2005discrete}
C.~Belta, V.~Isler, and G.~J. Pappas, ``Discrete abstractions for robot motion
  planning and control in polygonal environments,'' \emph{IEEE Trans. Robot.},
  vol.~21, no.~5, pp. 864--874, 2005.

\bibitem{rousseas2021harmonic}
P.~Rousseas, C.~Bechlioulis, and K.~J. Kyriakopoulos, ``Harmonic-based optimal
  motion planning in constrained workspaces using reinforcement learning,''
  \emph{IEEE Robot. Autom. Lett.}, vol.~6, no.~2, pp. 2005--2011, 2021.

\bibitem{goerzen2010survey}
C.~Goerzen, Z.~Kong, and B.~Mettler, ``A survey of motion planning algorithms
  from the perspective of autonomous uav guidance,'' \emph{J. Intell. Robot.
  Syst.}, vol.~57, pp. 65--100, 2010.

\bibitem{sarkar2009greedy}
R.~Sarkar, X.~Yin, J.~Gao, F.~Luo, and X.~D. Gu, ``Greedy routing with
  guaranteed delivery using {R}icci flows,'' in \emph{2009 International
  Conference on Information Processing in Sensor Networks}.\hskip 1em plus
  0.5em minus 0.4em\relax IEEE, 2009, pp. 121--132.

\bibitem{constantinou2020robot}
N.~Constantinou and S.~G. Loizou, ``Robot navigation on star worlds using a
  single-step navigation transformation,'' in \emph{2020 59th IEEE Conference
  on Decision and Control (CDC)}.\hskip 1em plus 0.5em minus 0.4em\relax IEEE,
  2020, pp. 1537--1542.

\bibitem{vlantis2018robot}
P.~Vlantis, C.~Vrohidis, C.~P. Bechlioulis, and K.~J. Kyriakopoulos, ``Robot
  navigation in complex workspaces using harmonic maps,'' in \emph{2018 IEEE
  International Conference on Robotics and Automation (ICRA)}.\hskip 1em plus
  0.5em minus 0.4em\relax IEEE, 2018, pp. 1726--1731.

\bibitem{gao2021conformal}
S.~Gao and N.~Bezzo, ``A conformal mapping-based framework for robot-to-robot
  and sim-to-real transfer learning,'' in \emph{2021 IEEE/RSJ International
  Conference on Intelligent Robots and Systems (IROS)}.\hskip 1em plus 0.5em
  minus 0.4em\relax IEEE, 2021, pp. 1289--1295.

\bibitem{notomista2018coverage}
G.~Notomista and M.~Egerstedt, ``Coverage control for wire-traversing robots,''
  in \emph{2018 IEEE International Conference on Robotics and Automation
  (ICRA)}.\hskip 1em plus 0.5em minus 0.4em\relax IEEE, 2018, pp. 5042--5047.

\bibitem{fan2022robot}
L.~Fan, J.~Liu, W.~Zhang, and P.~Xu, ``Robot navigation in complex workspaces
  using conformal navigation transformations,'' \emph{IEEE Robot. Autom.
  Lett.}, vol.~8, no.~1, pp. 192--199, 2022.

\bibitem{notomista2021safety}
G.~Notomista and M.~Saveriano, ``Safety of dynamical systems with multiple
  non-convex unsafe sets using control barrier functions,'' \emph{IEEE Control
  Syst. Lett.}, vol.~6, pp. 1136--1141, 2021.

\bibitem{ames2016control}
A.~D. Ames, X.~Xu, J.~W. Grizzle, and P.~Tabuada, ``Control barrier function
  based quadratic programs for safety critical systems,'' \emph{IEEE Trans.
  Autom. Control.}, vol.~62, no.~8, pp. 3861--3876, 2016.

\bibitem{lehto1973quasiconformal}
O.~Lehto, \emph{Quasiconformal mappings in the plane}.\hskip 1em plus 0.5em
  minus 0.4em\relax Springer-Verlag Berlin Heidelberg, 1973, vol. 126.

\bibitem{choi2020efficient}
G.~P.~T. Choi, ``Efficient conformal parameterization of multiply-connected
  surfaces using quasi-conformal theory,'' \emph{J. Sci. Comput.}, vol.~87,
  no.~3, pp. 1--19, 2021.

\bibitem{pinkall1993computing}
U.~Pinkall and K.~Polthier, ``Computing discrete minimal surfaces and their
  conjugates,'' \emph{Exp. Math.}, vol.~2, no.~1, pp. 15--36, 1993.

\bibitem{choi2015fast}
P.~T. Choi and L.~M. Lui, ``Fast disk conformal parameterization of
  simply-connected open surfaces,'' \emph{J. Sci. Comput.}, vol.~65, no.~3, pp.
  1065--1090, 2015.

\bibitem{lui2013texture}
L.~M. Lui, K.~C. Lam, T.~W. Wong, and X.~Gu, ``Texture map and video
  compression using {B}eltrami representation,'' \emph{SIAM J. Imaging Sci.},
  vol.~6, no.~4, pp. 1880--1902, 2013.

\bibitem{choi2015flash}
P.~T. Choi, K.~C. Lam, and L.~M. Lui, ``{FLASH}: Fast landmark aligned
  spherical harmonic parameterization for genus-0 closed brain surfaces,''
  \emph{SIAM J. Imaging Sci.}, vol.~8, no.~1, pp. 67--94, 2015.

\bibitem{choi2020parallelizable}
G.~P.~T. Choi, Y.~Leung-Liu, X.~Gu, and L.~M. Lui, ``Parallelizable global
  conformal parameterization of simply-connected surfaces via partial
  welding,'' \emph{SIAM J. Imaging Sci.}, vol.~13, no.~3, pp. 1049--1083, 2020.

\bibitem{marshall2007convergence}
D.~E. Marshall and S.~Rohde, ``Convergence of a variant of the zipper algorithm
  for conformal mapping,'' \emph{SIAM J. Numer. Anal.}, vol.~45, no.~6, pp.
  2577--2609, 2007.

\bibitem{wilson2020robotarium}
S.~Wilson, P.~Glotfelter, L.~Wang, S.~Mayya, G.~Notomista, M.~Mote, and
  M.~Egerstedt, ``The robotarium: Globally impactful opportunities, challenges,
  and lessons learned in remote-access, distributed control of multirobot
  systems,'' \emph{IEEE Control Syst. Mag.}, vol.~40, no.~1, pp. 26--44, 2020.

\bibitem{olfati2002near}
R.~Olfati-Saber, ``Near-identity diffeomorphisms and exponential/spl
  epsi/-tracking and/spl epsi/-stabilization of first-order nonholonomic se (2)
  vehicles,'' in \emph{Proceedings of the 2002 American Control Conference
  (IEEE cat. no. ch37301)}, vol.~6.\hskip 1em plus 0.5em minus 0.4em\relax
  IEEE, 2002, pp. 4690--4695.

\bibitem{khalil2015nonlinear}
H.~K. Khalil, \emph{Nonlinear control}.\hskip 1em plus 0.5em minus 0.4em\relax
  Pearson New York, 2015.

\bibitem{fliess1995flatness}
M.~Fliess, J.~L{\'e}vine, P.~Martin, and P.~Rouchon, ``Flatness and defect of
  non-linear systems: introductory theory and examples,'' \emph{Int. J.
  Control}, vol.~61, no.~6, pp. 1327--1361, 1995.

\end{thebibliography}
